%% file: main.tex
\documentclass[11pt]{article}
\PassOptionsToPackage{table}{xcolor}

\usepackage[preprint]{acl}

\usepackage{times}
\usepackage{latexsym}

\usepackage[T1]{fontenc}

\usepackage[utf8]{inputenc}

\usepackage{microtype}

\usepackage{inconsolata}

\usepackage{graphicx}
\usepackage{amsmath,amssymb}
\usepackage{float}
\usepackage{booktabs,array,makecell}
\definecolor{navy_blue}{RGB}{0,47,167}
\title{Kongzi: A Historical Large Language Model with Fact Enhancement}

\author{
    Jiashu Yang\textsuperscript{1,*}\textsuperscript{\ensuremath{\dagger}}, 
    Ningning Wang\textsuperscript{1,*}, 
    Yian Zhao\textsuperscript{3}, 
    Chaoran Feng\textsuperscript{3}, 
    Junjia Du\textsuperscript{2}, \\
    \vspace{0.5em}
    \textbf{Hao Pang}\textsuperscript{1}, 
    \textbf{Miao Yu}\textsuperscript{1},
    \textbf{Xuxin Cheng}\textsuperscript{3}\textsuperscript{\ensuremath{\dagger}}\\
    \textsuperscript{1}Dalian University of Technology\\
    \textsuperscript{2}Nanyang Technological University\\
    \textsuperscript{3}Peking University\\
    \texttt{jiashuyang827@gmail.com, chengxx.pku@gmail.com}
}

\begin{document}
\maketitle
\begingroup
\renewcommand\thefootnote{}
\footnote{\textsuperscript{*}Equal contribution. \textsuperscript{\ensuremath{\dagger}}Corresponding authors.}
\endgroup
\input{sections/abstract}
\input{sections/introducion}
\input{sections/related}
\input{sections/method}
\input{sections/experiment}
\input{sections/conclusion}

\bibliography{custom}

\appendix
\clearpage

\onecolumn  
\section{Training Configurations}  
\input{table/table3.tex}
\clearpage

\section{Test Samples}  
\input{image/kong11}
\input{image/kong12.tex}
\input{image/kong13.tex}
\input{image/kong21.tex}
\input{image/kong22.tex}
\input{image/kong23.tex}
\input{image/kong30.tex}
\input{image/kong31.tex}
\input{image/kong32.tex}
\input{image/kong33.tex}

\end{document}

%% file: sections/abstract.tex
\begin{abstract}

Large Language Models (LLMs) can exhibit factual inaccuracies during long-chain reasoning or under misleading prompts, limiting their reliability in historical analysis. We propose \textbf{Kongzi}, a language model for historical understanding and reasoning that combines continued pre-training on historical corpora, supervised fine-tuning, and fact-enhanced reinforcement learning. Its factuality reward compares entities and relations extracted from generated responses with historical references through knowledge-graph similarity. We evaluate 0.5B and 7B models on historical question answering and counterfactual reasoning, using reported ratings from three LLM judges, and on CMMLU-Chinese-History. Kongzi-7B receives the highest historical-accuracy ratings among the compared models under all three judges. Reinforcement learning improves the mean answer scores of both model sizes and their CMMLU-Chinese-History accuracy, although Kongzi-7B remains below the strongest reported baselines on that benchmark. These descriptive results support further study of factuality rewards for historical language models.

\end{abstract}

%% file: sections/introducion.tex
\section{Introduction}

Recent advancements in large language models have achieved significant breakthroughs ranging from text generation to complex reasoning \citep{plaat2024reasoninglargelanguagemodels,deepseekai2025deepseekr1incentivizingreasoningcapability}. 
Although LLMs have achieved strong performance across general domains, their application to historical tasks remains a challenge. 
Historical text generation demands both factual accuracy and the ability to infer complex relationships from fragmented evidence~\citep{cao2024tonggumasteringclassicalchinese}. However, existing models often suffer from hallucinations (generating factually inaccurate content), particularly under the misleading prompts~\citep{Huang_2025,chen2024inside}, which limits their reliability in real-world historical analysis.

\input{image/motivation}

Existing methods to enhance LLMs include reinforcement learning from human feedback (RLHF) \citep{ouyang2022traininglanguagemodelsfollow} and chain-of-thought (CoT) fine-tuning\citep{chen2024learning,kim2023the}. 
These approaches primarily target general-purpose tasks, focusing on aligning model outputs with human preferences or emulating human-like reasoning processes. However, they exhibit notable limitations when applied to domains with high factual precision requirements. 
For instance, RLHF often lacks domain-specific factual constraints, resulting in outputs that may be fluent and preference-aligned, but lack logical consistency with verified facts. 
Similarly, while CoT fine-tuning enhances reasoning ability, it does not inherently prevent the generation of factually inaccurate content. 
These limitations underscore the need for a domain-adaptive training framework for the historical domain---where factual reliability and deep, context-aware reasoning are essential.

Therefore, to address this issue, we propose a reasoning-oriented large language model specifically tailored for historical analysis tasks---\textbf{Kongzi}. By enhancing the model's understanding of historical facts, Kongzi is able to better analyze and answer historical questions, while achieving superior factual accuracy and consistency.

Specifically, we first collect domain-specific corpora related to history and conduct continued pre-training on top of a base model. We then perform instruction-following fine-tuning via a cold-start strategy using carefully constructed instruction data. Finally, we introduce a novel reward function: by building a graph of relevant information (entities and relations) and measuring its similarity to the ground truth, we use this as a reward signal to minimize the discrepancy between model output and factual references, thus improving the factual awareness of the model. We conduct comprehensive evaluations on public benchmarks as well as expert assessments. The reported results allow comparisons of factual accuracy and reasoning quality across the evaluated models.

In summary, our contributions can be stated as follows:

\begin{itemize}
\item
Continued Pre-training: We curated a diverse corpus of historical texts and conducted continued pre-training to adapt the base model to the historical domain. This step ensures enhanced domain-specific knowledge for smaller models.  

\item
CoT Data Generation: We classified initial data into factual statements and reasoning tasks, generating high-quality CoT data to enhance the model's reasoning ability through two-stage supervised fine-tuning (SFT). This approach enables the model to perform deep contextual reasoning, a critical capability for historical applications. 

\item
Fact-Aware RL: We proposed a novel reward function that incorporates factual rewards into the reinforcement learning framework, ensuring that the model generates content aligned with historical facts.
\end{itemize}

%% file: image/motivation.tex
\begin{figure}[t]
    \centering
    \includegraphics[width=\columnwidth]{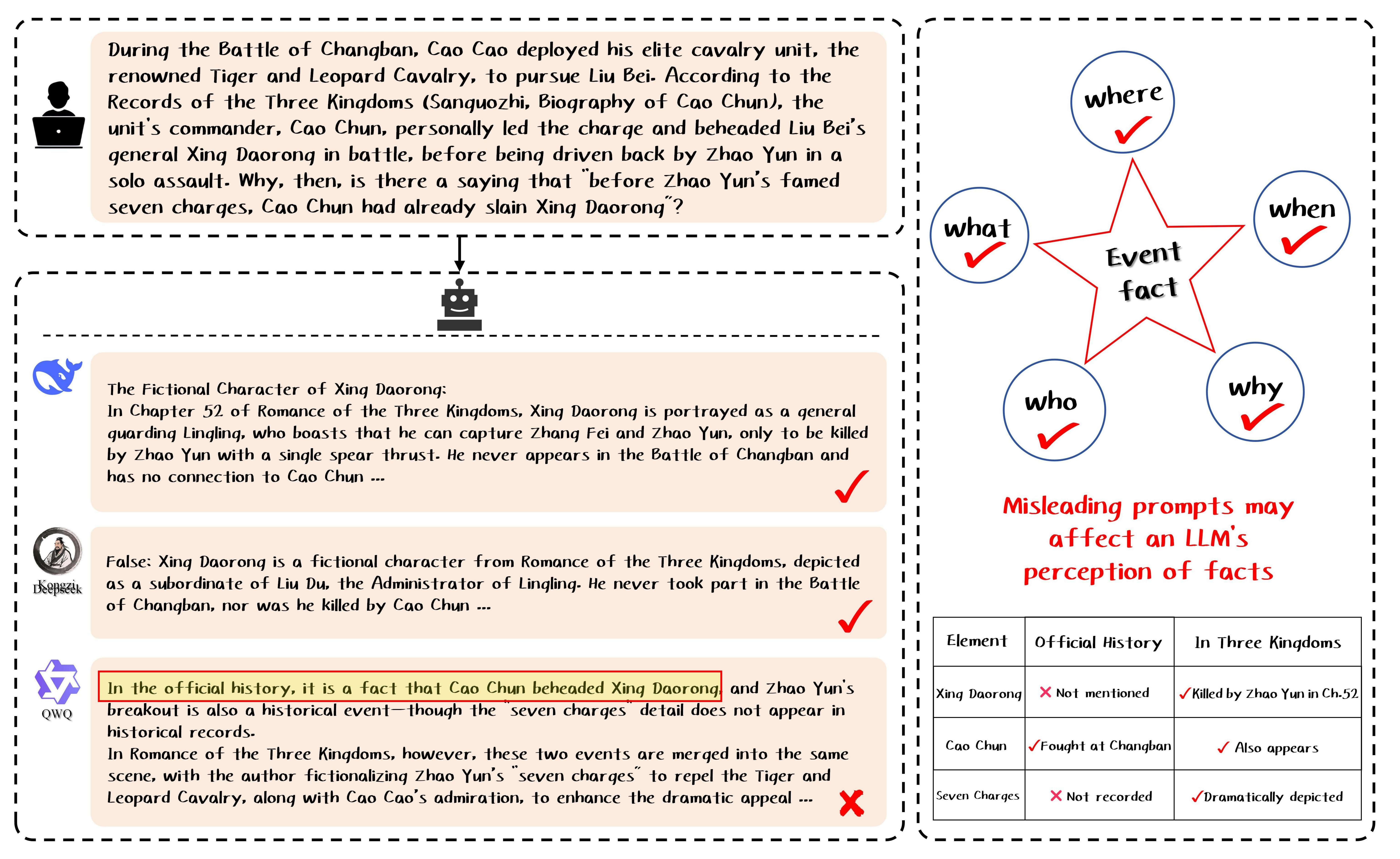}
    \vspace{+0.02in}
    \caption{
     Comparative analysis of different LLM's responses to misleading historical fact-checking queries.
    }
    \label{fig:motivation}
\end{figure}

%% file: sections/related.tex
\section{Related Work}

\subsection{Reasoning in Large Language Models}
In recent years, large reasoning models have shown incredible capabilities in a series of complex tasks, enabling them to solve complex tasks by decomposing problems into logical steps. Early reasoning approaches primarily relied on prompt engineering and few-shot learning \citep{yu2023betterchainofthoughtpromptingstrategies}, guiding models to generate intermediate reasoning steps through carefully designed prompts. While these methods excelled in handling simple tasks, they often struggled with more complex problems due to their dependence on static prompts and limited context. Recent research has explored dynamic reasoning frameworks, such as self-consistent decoding \citep{wang2023selfconsistencyimproveschainthought} and scratchpad prompting \citep{nye2021workscratchpadsintermediatecomputation}, which allow models to iteratively refine their reasoning processes.

\textbf{Chain-of-Thought fine-tuning} has emerged as a crucial technique for enhancing the reasoning capabilities of LLMs \citep{kim2023the}. By generating intermediate reasoning steps, CoT enables models to break down complex problems into simpler subtasks, resulting in more accurate and interpretable solutions. For example, zero-shot CoT demonstrated that LLMs can generate reasoning chains even without explicit fine-tuning, while CoT distillation \citep{wadhwa2024investigatingmysteriescotaugmenteddistillation} showed that high-quality CoT data can significantly boost model performance~\citep{yang2025reasonfluxhierarchicalllmreasoning}.

\input{image/pipeline}

\textbf{Reinforcement Learning for Reasoning}. Process-based reward models guide the model to perform structured problem-solving, while RL methods optimize performance through feedback-driven learning. Although these methods have achieved incremental gains, they still fall short of the general reasoning capabilities demonstrated by the o1. Recently, the DeepSeek-R1 model \citep{deepseekai2025deepseekr1incentivizingreasoningcapability} has emerged as a competitive alternative, utilizing RL with the Group Relative Policy Optimization \citep{deepseekai2025deepseekr1incentivizingreasoningcapability} algorithm, but the reward function it uses is only suitable for mathematical and coding tasks. Recent studies~\citep{rafailov2024directpreferenceoptimizationlanguage,schulman2017proximalpolicyoptimizationalgorithms,zhong2025dpomeetspporeinforced,ramesh2024group} have explored fact-aware RL where models are rewarded for generating factually accurate content, but these approaches typically focus on single-turn tasks and do not address the challenges of lengthy historical narratives. Our work introduces a novel fact-aware RL framework that evaluates the accuracy of each generated segment, ensuring consistent factual correctness throughout the generation process.

\subsection{Chinese History in Large Language Models}

Recent work has begun to explore large pretrained language models (LLMs) for multimodal understanding and generation within Chinese cultural contexts. For example, \citep{li2024foodieqa} introduce the FoodieQA dataset, and \citep{zhao2025fuxibenchmarkevaluatinglanguage} propose the Fuxi, a systematic assessment framework that considers Chinese cultural authenticity.

In modeling and question--answering over Chinese historical knowledge, several studies have combined LLMs with structured knowledge bases to enhance factual accuracy and response quality. Despite advances in factual retrieval and knowledge integration, systematic evaluation of LLM temporal reasoning and multi-entity alignment remains scarce. The CTM benchmark \citep{wang2025benchmarkingtemporalreasoningalignment} was proposed to address this gap, which constructs QA pairs and Timeline Ito Games that rigorously assess the abilities of models in cross-dynastic entity ordering, event alignment, and reasoning. In our work, we build on these insights by designing a history-enhanced reward function for reinforcement learning, and investigate how external knowledge injection can further optimize the temporal reasoning and alignment capabilities of models.

%% file: image/pipeline.tex
\begin{figure*}[t]
    \centering
    \includegraphics[width=\textwidth]{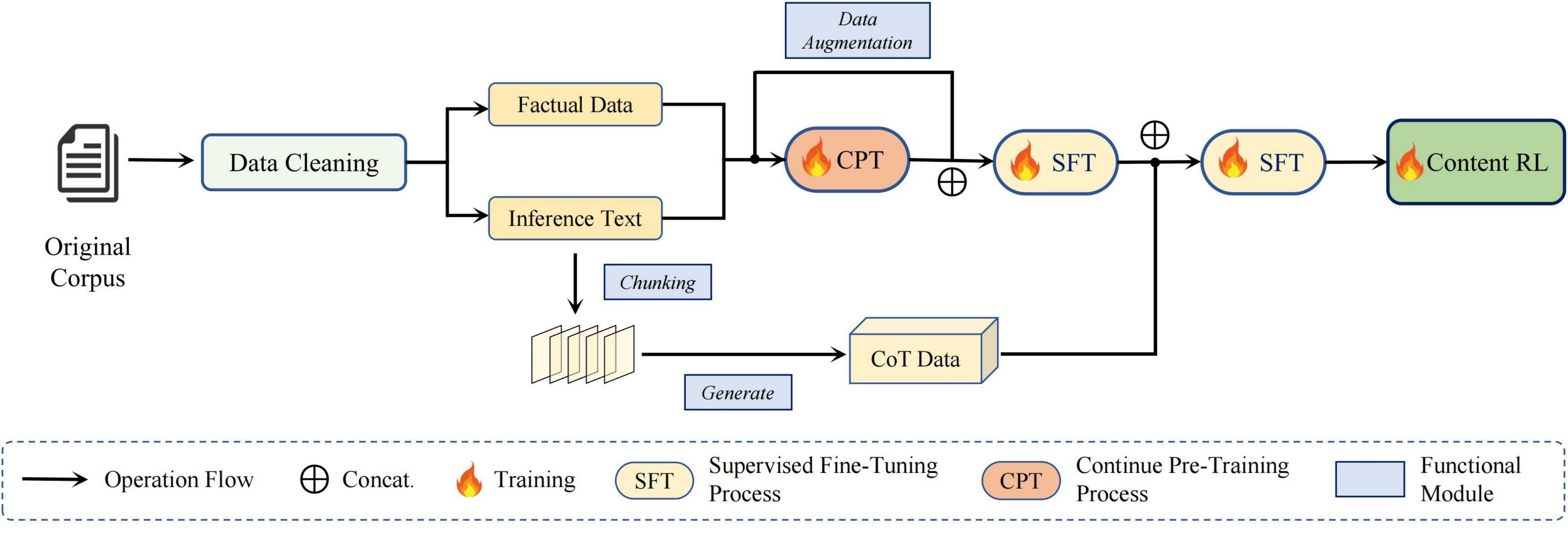}
    \vspace{+0.02in}
    \caption{
    \textbf{Overview of our pipeline.} In terms of data processing, we clean the original corpus and separate it into factual data and inference text, where factual data undergoes augmentation while inference text is chunked for generating high-quality CoT data. For model training, we first conduct CPT with basic data, followed by a two-stage SFT to develop both fundamental QA capabilities and CoT reasoning abilities. Finally, Content RL is employed to optimize output quality through reinforcement learning.
    }
\label{fig:pipeline}
\end{figure*}

%% file: sections/method.tex
\section{Method}

To improve the factual accuracy of large models in the historical domain, we first collected and cleaned vertical domain data, then performed incremental pre-training on the pre-trained model to enhance its understanding of relevant information. Next, we conducted cold-start instruction fine-tuning to optimize the response structure, and finally applied a fact-enhanced reinforcement learning algorithm for further training. This training pipeline is designed to improve factual consistency and reasoning performance in historical tasks.

\subsection{Data acquisition and processing}
\label{sec:3.1}
Considering the continuity of ancient Chinese history and the completeness of its historical records, we selected it as the primary source of historical text data. With the assistance of domain experts, we effectively collected approximately 0.2 billion tokens of high-quality Chinese historical texts.
Continued pre-training data:
Data cleaning such as reducing noise and deduplication of the original corpus of 0.2B tokens
SFT data:

(1) Classification: Historical corpora can be objectively categorised into two types: factual statements and contextual reasoning texts. The latter are used for subsequent data generation purposes. We employ a small-parameter LLM-assisted workflow for this separation. Specifically, when determining contextual information, the model segments text into multiple sentences based on punctuation marks, assigning each sentence a unique identifier. The incorporation of adjacent sentences as auxiliary context is necessitated by the potential for isolated sentences to lack semantic coherence and consequently lead to misjudgement. The remaining content is designated as factual information.

(2) Generation of Primary SFT Data: To enhance the model's fundamental question-answering capability, we employ a random sampling method to construct simple question-and-answer pairs via LLM.

(3) Generation of Secondary SFT Data: The incorporation of chain-of-thought (CoT) data is imperative to facilitate the model's reasoning capacity. By leveraging the classified contextual reasoning texts, we guide the reasoning model to perform step-by-step analysis on each sentence and its contextual content. The output of this process follows a standardised CoT format, systematically constructing reasoning chain datasets.

\subsection{Initial training strategy for the model}
\label{sec:3.2}
In the Continued pre-training (CPT) stage, the model has already obtained relevant historical information, but its command question-answering ability has not yet reached normal standards. To this end, we preliminarily optimize the pre-trained model through supervised command fine-tuning, so that it has preliminary command understanding and execution capabilities. To further enable the model to acquire reasoning capabilities, we introduce high-quality Chain-of-Thought data, and through the SFT method, enable the model to generate text according to the preset Chain-of-Thought template, thereby constructing a model architecture with preliminary reasoning capabilities, enabling it to have reasoning capabilities.

\subsection{Reinforcement Learning Training}
\label{sec:3.3}
Previous studies aimed at enhancing the reasoning capabilities of language models have mainly adopted three strategies: first, using reward mechanisms to encourage models to output detailed reasoning processes; second, extending the model's reasoning time (i.e., increasing the number of output tokens); and third, using Retrieval-Augmented Generation (RAG) to address hallucinations caused by insufficient information through external knowledge bases. However, all three strategies have significant limitations: when the generated text becomes lengthy, the factual accuracy of the middle content becomes difficult to guarantee, making hallucinations more likely. 
\input{image/reward}
Specifically, while language models can generate accurate short texts with clear prompts, they are more prone to factual errors or hallucinations in long-form generation, especially when context is limited or noisy data is introduced.

To address this, we propose a reinforcement learning-based optimization strategy. We introduce the Historical Fact-enhanced Reward Module (HFRM), which focuses on factual consistency by extracting entities and relations from generated text to build a candidate knowledge graph. This graph is compared with a ground truth graph, and their semantic similarity yields a factuality score that serves as part of the reward. This reward provides a structured signal for aligning generated historical content with the reference facts.

\input{image/score}

Let
\begin{align}
X &= \frac{\sum_{i=1}^{G} \sum_{t=1}^{\|o_i\|} \mathcal{M}(r_{ratio}) \cdot \hat{A}_{i,t}}{G \sum_{i=1}^{G} \|o_i\|}, \\
\mathcal{M}(r) &= \min \left[ r, \operatorname{CLIP}(r, 1-\varepsilon, 1+\varepsilon) \right]
\end{align}
where:
\begin{itemize}
    \item \( G \) is the size of the action group sampled for each state.
    \item \( o_i \) is the output sequence of the \( i \)-th action group.
    \item \( \|o_i\| \) is the length of the output sequence of the \( i \)-th action group.
    \item \( r_{\mathrm{ratio}} \) is the probability ratio of the current policy to the old policy.
    \item \( \varepsilon \) is the clipping parameter for the probability ratio.
    \item \( \hat{A}_{i,t} \) is the relative advantage of the \( i \)-th action group at time step \( t \).
\end{itemize}

The objective function of Group Relative Policy Optimization (GRPO) can be expressed as:
\begin{equation}
\mathcal{J}^{\mathrm{GRPO}}(\theta) = \mathbb{E}[X] - \varepsilon_{KL}
\end{equation}
where:
\begin{itemize}
    \item $\varepsilon_{KL}$ is the KL divergence regularization term, used to constrain the difference between the new and old policies.
\end{itemize}
This objective function optimizes the policy model's parameters \( \theta \) by maximizing the expectation of \( X \), while introducing a KL divergence regularization term.

To enhance the factual consistency and semantic reliability of language models in knowledge-intensive tasks, we propose a Factuality-Oriented Reward Mechanism for Historical Entities. This mechanism aims to reinforce the factual accuracy and relational completeness of long-form text generation. It is built upon the traditional Group Relative Policy Optimization (GRPO) framework, with the introduction of a novel component---Historical Fact-enhanced Reward Module (HFRM)---designed to more precisely capture the alignment between generated text and historical facts.

Specifically, we first perform information extraction on the generated text to identify entities such as historical figures, locations, and time points, as well as the relationships among them. Based on these, we construct a candidate knowledge graph corresponding to the generated content. This graph is then aligned with a ground truth knowledge graph curated by human experts. By calculating the semantic similarity between the two graphs---e.g., via knowledge graph embedding techniques---we derive a factuality score, which serves as one of the reward signals in the reinforcement learning process.

In terms of reward design, we adopt a fine-grained strategy: when the model correctly generates entities and relations that match the ground truth, a positive reward is given to encourage accurate learning and expression of historical facts. Conversely, if the generated output contains missing entities, incorrect recognition, distorted relations, or semantic inconsistencies (such as reversed relations or temporal logic errors), a negative reward is applied to suppress the generation of hallucinated or incoherent information.

By explicitly modeling factuality at the entity level, this mechanism aims to reduce factual errors in complex, multi-sentence historical texts.

%% file: image/reward.tex
\begin{figure*}[t]
    \centering
    \includegraphics[width=\textwidth]{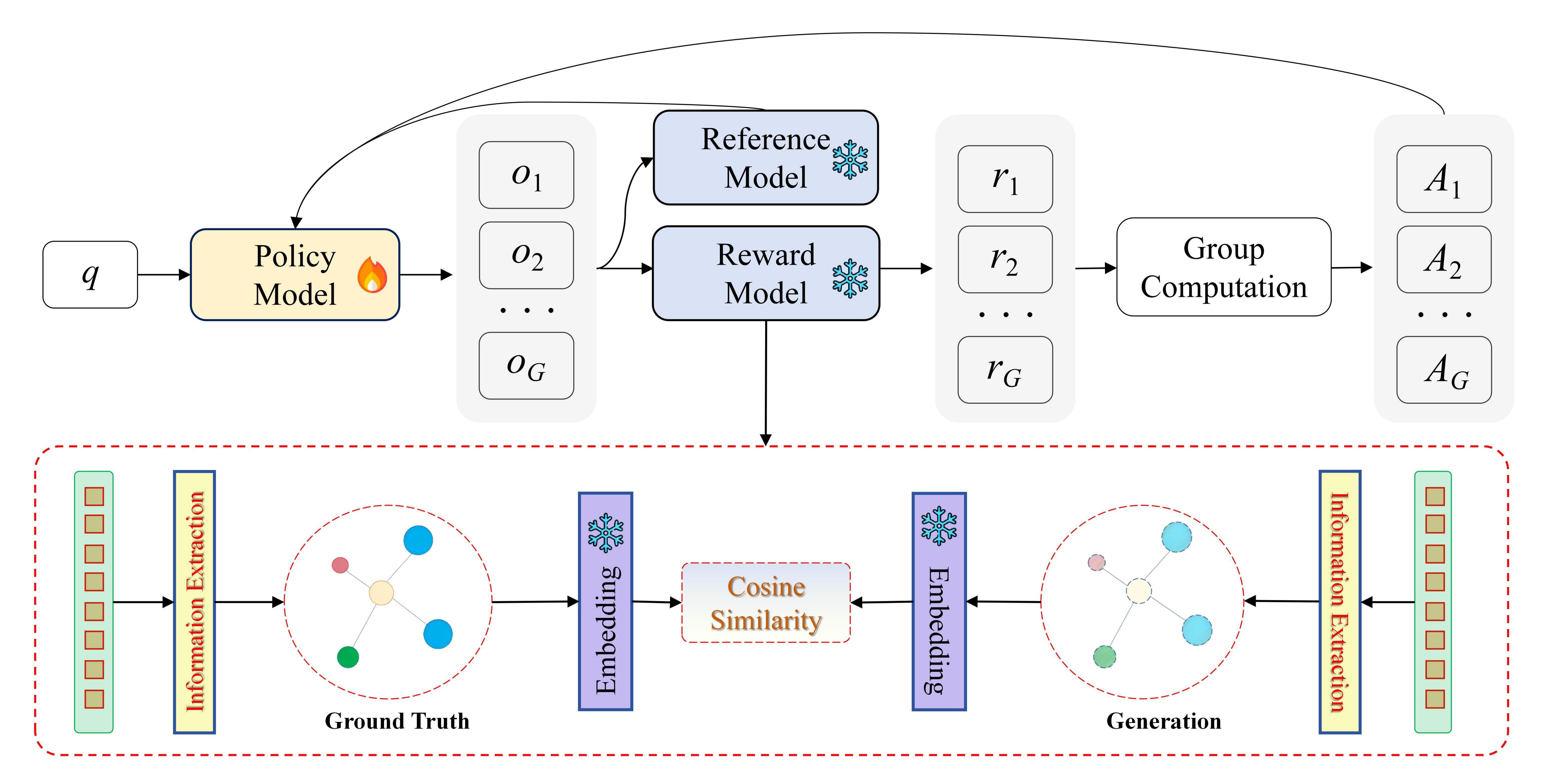}
    \vspace{+0.02in}
    \caption{
     The HFRM module constructs a knowledge graph by extracting entities and relations from the model outputs, which serves as the foundation for information representation and computation. The semantic similarity between this graph and the ground truth is used as the reward signal.
    }
\label{fig:reward}
\end{figure*}

%% file: image/score.tex
\begin{figure*}[t]
    \centering
    \includegraphics[width=\textwidth]{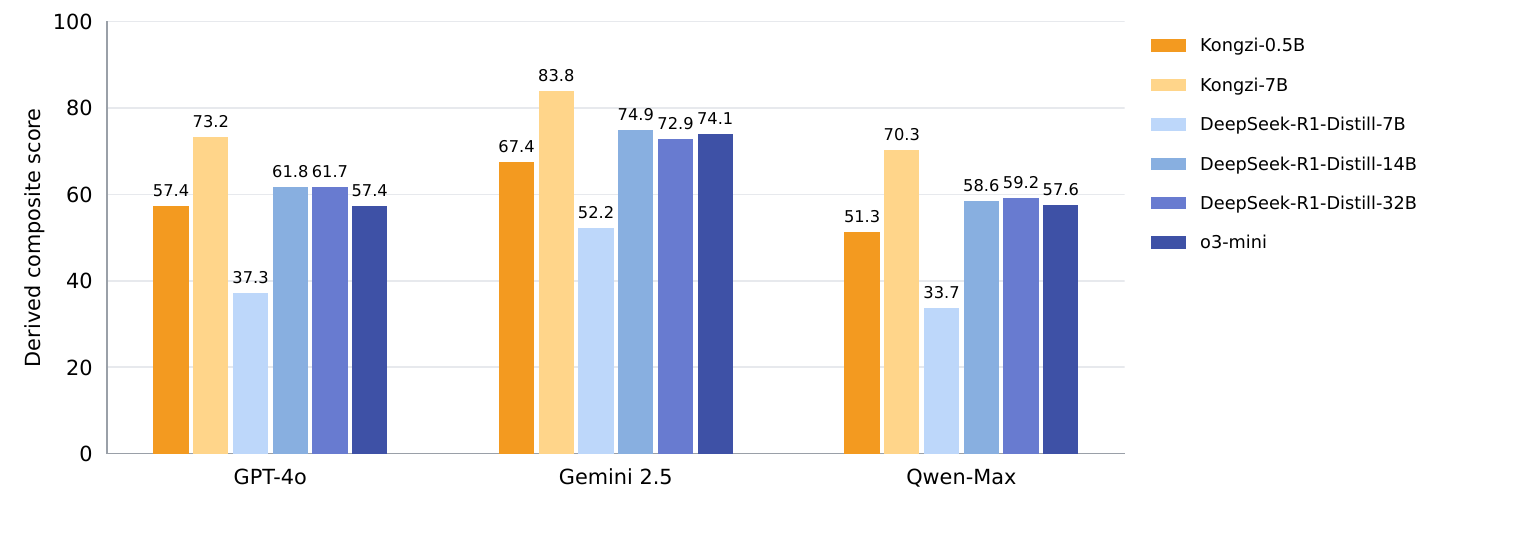}
    \caption{\textbf{Composite scores recomputed from the reported aggregate data.} For models with all required components, the score is $0.8A+0.1T+0.1W$, where $A$ is the arithmetic mean of the three displayed answer-dimension scores, $T$ is Think Score, and $W$ is the reported ratio favoring the model over Deepseek-r1 on the 0--100 scale. Means are used before rounding. Deepseek-r1 and QwQ are omitted from this composite plot because their reported Ratio and Think Score, respectively, are unavailable; their available component scores remain in Table~\ref{tab:comparison}. These derived composites do not represent a new evaluation run.}
    \label{fig:score}
\end{figure*}

%% file: sections/experiment.tex
\section{Experiments}
\subsection{Datasets and Evaluation Protocols}
Given the relatively complete and continuous nature of ancient Chinese history, we have collected nearly 200 million tokens of Chinese historical and cultural annotated text corpus. This includes, but is not limited to, publicly available sources such as the Twenty-Four Histories. The corpus spans thousands of years of historical periods, events, and figures, ensuring a broad coverage throughout ancient Chinese history.

To facilitate targeted model training and obtain higher-quality data, we have objectively categorized the corpus into two types:
\begin{enumerate}
    \item \textbf{Factual statement data:} Mainly contains objective descriptions of historical events, such as dates, places, and relationships between people.
    \item \textbf{Contextual reasoning text:} Contains inferable value through contextual relationships, metaphors, and proverbs, such as analysis of historical causal relationships or discussion of the meaning of certain events.
\end{enumerate}

\subsection{Data Processing \& Generating}
Since most of the original texts are written in Classical Chinese, directly using them for continued pre-training may lead to semantic ambiguities in vernacular Chinese contexts. Therefore, after basic cleaning and preprocessing of the original data, we randomly sampled 700,000 entries and employed a large language model (LLM) to automatically generate question--answer pairs in both Classical and vernacular Chinese styles, thereby enhancing the training data for the subsequent supervised fine-tuning (SFT) stage.

For the factual historical data, we randomly selected and filtered 3,000 samples. The entities and relations in each sentence were extracted and then verified by domain experts.

For the contextual reasoning texts, we used an LLM to automatically identify 1,300 representative historical events and questions. Among them, 1,000 samples were constructed into structured instruction-tuning training data in Chain-of-Thought (CoT) format using Deepseek-R1, while the remaining 300 samples were used as a test set.

\input{table/table2}
\subsection{Base Model Training}
We used Qwen2.5 \citep{qwen2025qwen25technicalreport} with 0.5B and 7B parameters as the base model for our experiments.

\subsubsection{Continued PreTraining}
We performed continued pre-training on the 0.5B and 7B models using cleaned classical Chinese data on the NVIDIA A6000 GPU for 1 and 5 days, respectively.

\subsubsection{Supervised Fine-Tuning}
After continued pre-training, we conducted the first step of SFT on 1 million augmented data using NVIDIA A6000 and eight NVIDIA GeForce RTX 3090 GPUs. To enhance the model's ability to answer historical questions while improving its general conversational capabilities, we used a combination of general datasets and our augmented dataset for SFT to achieve human alignment.

In the first step of the SFT process, we adopted a two-stage training strategy to balance general conversational ability and historical question-answering ability:
\begin{enumerate}
    \item In the first stage, we performed SFT training using a general SFT dataset and half of our augmented dataset with block-level random mixing.
    \item In the second stage, we used the other half of the augmented dataset for SFT.
\end{enumerate}
\input{table/cmmlu.tex}

In the second-stage SFT, we fine-tuned the model on 1,000 Chain-of-Thought (CoT) examples for structured instruction following. This cold-start step trains the model to follow long-form CoT trajectories and provides an initialization for the subsequent RL phase.

\subsection{Reinforcement Learning Training}
After completing continued pre-training and SFT, we further optimized the model's performance through RL using 3000 data points.

\subsubsection{Reinforcement Learning Framework}
We adopt GRPO as the reinforcement learning framework, incorporating a factuality reward to encourage factual consistency during generation. In this reward function, we compute the cosine similarity between the knowledge graphs of the ground truth and the model output as the reward signal.

Specifically, we designed the following reward functions to assess the model's output:

\begin{itemize}
    \item Formula:
    \[
    ER = \cos \left( f(\mathcal{G}_{\text{model}}),\ f(\mathcal{G}_{\text{gold}}) \right)
    \]
    where \(\mathcal{G}_{\text{model}}\) and \(\mathcal{G}_{\text{gold}}\) denote the knowledge graphs extracted from the model output and the ground-truth, respectively, and \(f(\cdot)\) is a graph embedding function that maps a knowledge graph to a dense vector representation.
\end{itemize}

\begin{enumerate}
    \item \textbf{Format Reward:}
    \begin{itemize}
        \item Check for format markers like \texttt{\textless\textbar begin\_of\_thought\textbar\textgreater}.
        \item Reward: 1.0 for correct format, 0.0 otherwise.
    \end{itemize}

    \item \textbf{Repetition Penalty Reward:}
    \begin{itemize}
        \item Penalize repeated n-grams with a negative value based on repetition ratio.
        \item Formula:
        \[
        {RP} = {RR} \times {MPV}
        \]
    \end{itemize}
\end{enumerate}
\subsubsection{Integration of Reward Functions}
We combined multiple reward functions with weighted aggregation. The final reward is calculated as follows:

\begin{equation}
    R_{\text{final}} = w_1 R_{\text{cos}} + w_2 R_{\text{log}} + w_3 R_{\text{fmt}} + w_4 R_{\text{rep}}
\end{equation}

where:
\begin{itemize}
    \item $R_{\text{cos}}$: Factual reward
    \item $R_{\text{log}}$: Logical coherence reward
    \item $R_{\text{fmt}}$: Format reward
    \item $R_{\text{rep}}$: Repetition penalty reward
    \item $w_1, w_2, w_3, w_4$: Corresponding weights, where \(\sum_{i=1}^{4} w_i = 1\)
\end{itemize}

\subsection{Results Evaluation}
\input{table/table1}

We report results on CMMLU-Chinese-History and our self-constructed test set of 300 samples. For the constructed test set, the reported LLM ratings use GPT-4o \citep{openai2024gpt4technicalreport}, Gemini-2.5, and Qwen-Max. Using multiple judges provides complementary assessments given their different knowledge and preferences \citep{gu2025surveyllmasajudge}. A manual assessment involving twelve volunteers reported 72.13\% support for Kongzi-7B; individual votes and their aggregation are not provided in the current materials.

The reported rating dimensions are \textit{Think Score}, which assesses the reasoning process or strategy; \textit{Historical Accuracy}, which assesses statements about historical events; \textit{Logical Reasoning}, which assesses the coherence of the reasoning; and \textit{Problem Solving}, which assesses how well the response addresses the question. \textit{Mean Answer Score} is derived from the displayed aggregates of the latter three dimensions as specified in Table~\ref{tab:comparison}; it is not a new evaluation run. The \textit{$>$DeepSeek-R1 Ratio} retains the reported percentage favoring each model over DeepSeek-R1.

Figure~\ref{fig:score} displays composite scores for the six models with complete inputs, using the weighting specified in its caption. DeepSeek-R1 and QwQ remain in Table~\ref{tab:comparison}; their composite scores are omitted because a required input is missing.

In Table~\ref{tab:comparison}, Kongzi-7B receives the highest Historical Accuracy ratings among the compared models under all three judges. Its advantage does not extend to every dimension: Gemini-2.5 assigns O3-mini a higher Logical Reasoning rating, and Qwen-Max assigns O3-mini a higher Problem Solving rating. Kongzi-0.5B exceeds DeepSeek-R1-distill-7B in the displayed composite under all three judges, while comparisons with larger models vary by judge and metric.

Table~\ref{tab:rl-improve} compares the SFT and RL versions. Mean Answer Score increases for both model sizes under all three judges, as do their Historical Accuracy ratings. The Qwen-Max Problem Solving rating for Kongzi-0.5B remains unchanged at 44.52.

On CMMLU-Chinese-History (Table~\ref{tab:cmmlu_kongzi}), accuracy increases from 59.37\% to 71.21\% for Kongzi-0.5B and from 87.56\% to 90.37\% for Kongzi-7B after RL. Kongzi-7B therefore improves over its SFT version, but its 90.37\% accuracy remains below the 92.88\% reported for DeepSeek-R1 and Qwen3-235B-a33B.

%% file: table/table2.tex
\begin{table*}[htbp]
\fontsize{9}{11}\selectfont 
\vspace{+0.2in}
\centering
\renewcommand{\arraystretch}{1.2}
\setlength{\tabcolsep}{6pt}
\rowcolors{3}{white}{gray!5}

\resizebox{\textwidth}{!}{
\begin{tabular}{
    >{\centering\arraybackslash}m{3.2cm}
    *{5}{>{\centering\arraybackslash}m{1.8cm}}
    >{\centering\arraybackslash}m{2.0cm}
}
\rowcolor{gray!20}
\toprule
\textbf{Model} & 
\makecell{\textbf{Think}\\ \textbf{Score}} & 
\makecell{\textbf{Mean}\\ \textbf{Answer}\\ \textbf{Score}} & 
\makecell{\textbf{Historical}\\ \textbf{Accuracy}} & 
\makecell{\textbf{Logical}\\ \textbf{Reasoning}} & 
\makecell{\textbf{Problem}\\ \textbf{Solving}} & 
\makecell{\textbf{$>$Deepseek-r1}\\ \textbf{Ratio}} \\
\toprule

\rowcolor{gray!10}
\multicolumn{7}{l}{\textbf{\textcolor{navy_blue}{GPT-4o}}} \\ 
\midrule
Kongzi-0.5B-SFT & 50.42 & 45.68 & 33.87 & 56.59 & 46.59 & 52.18\% \\
Kongzi-7B-SFT & 60.46 & 64.27 & 50.97 & 76.96 & 64.88 & 85.29\% \\
\textbf{Kongzi-0.5B-RL} & \textbf{60.42}\textsubscript{(+10.0)} & \textbf{56.39}\textsubscript{(+10.7)} & \textbf{53.34}\textsubscript{(+19.5)} & \textbf{62.41}\textsubscript{(+5.8)} & \textbf{53.43}\textsubscript{(+6.8)} & \textbf{62.28}\%\textsubscript{(+10.1)} \\
\textbf{Kongzi-7B-RL} & \textbf{72.34}\textsubscript{(+11.9)} & \textbf{70.64}\textsubscript{(+6.4)} & \textbf{58.45}\textsubscript{(+7.5)} & \textbf{82.13}\textsubscript{(+5.2)} & \textbf{71.34}\textsubscript{(+6.5)} & \textbf{94.32}\%\textsubscript{(+9.0)} \\

\midrule
\rowcolor{gray!10}
\multicolumn{7}{l}{\textbf{\textcolor{navy_blue}{Gemini-2.5}}} \\ 
\midrule
Kongzi-0.5B-SFT & 70.89 & 61.78 & 54.61 & 64.38 & 66.36 & 42.17\% \\
Kongzi-7B-SFT & 80.13 & 74.72 & 68.39 & 76.04 & 79.72 & 57.83\% \\
\textbf{Kongzi-0.5B-RL} & \textbf{77.91}\textsubscript{(+7.0)} & \textbf{68.49}\textsubscript{(+6.7)} & \textbf{61.13}\textsubscript{(+6.5)} & \textbf{69.88}\textsubscript{(+5.5)} & \textbf{74.45}\textsubscript{(+8.1)} & \textbf{48.17}\%\textsubscript{(+6.0)} \\
\textbf{Kongzi-7B-RL} & \textbf{89.24}\textsubscript{(+9.1)} & \textbf{85.10}\textsubscript{(+10.4)} & \textbf{80.45}\textsubscript{(+12.1)} & \textbf{85.43}\textsubscript{(+9.4)} & \textbf{89.42}\textsubscript{(+9.7)} & \textbf{68.21}\%\textsubscript{(+10.4)} \\

\midrule
\rowcolor{gray!10}
\multicolumn{7}{l}{\textbf{\textcolor{navy_blue}{Qwen-Max}}} \\ 
\midrule
Kongzi-0.5B-SFT & 54.40 & 44.32 & 32.12 & 56.31 & 44.52 & 36.17\% \\
Kongzi-7B-SFT & 64.46 & 61.52 & 46.59 & 78.43 & 59.54 & 79.49\% \\
\textbf{Kongzi-0.5B-RL} & \textbf{60.40}\textsubscript{(+6.0)} & \textbf{50.49}\textsubscript{(+6.2)} & \textbf{39.82}\textsubscript{(+7.7)} & \textbf{67.12}\textsubscript{(+10.8)} & \textbf{44.52}\textsubscript{(+0.0)} & \textbf{49.18}\%\textsubscript{(+13.0)} \\
\textbf{Kongzi-7B-RL} & \textbf{73.31}\textsubscript{(+8.9)} & \textbf{67.82}\textsubscript{(+6.3)} & \textbf{56.36}\textsubscript{(+9.8)} & \textbf{84.78}\textsubscript{(+6.4)} & \textbf{62.31}\textsubscript{(+2.8)} & \textbf{87.21}\%\textsubscript{(+7.7)} \\

\bottomrule
\end{tabular}
}

\vspace{0.2cm}
{\footnotesize
\caption{\textbf{Improvement of Kongzi model performance in various metrics by RL training.} Mean Answer Score and its changes are recomputed from the displayed Historical Accuracy, Logical Reasoning, and Problem Solving scores. Subscripts give RL-minus-SFT changes; ratio changes are in percentage points. Bold identifies RL rows. Other scores and ratios are retained as reported.}
\label{tab:rl-improve}
}
\end{table*}

%% file: table/cmmlu.tex
\begin{table}[t]
\renewcommand{\arraystretch}{1.2}
\setlength{\tabcolsep}{4pt}
\rowcolors{4}{white}{gray!5}

\small
\centering

\resizebox{0.95\linewidth}{!}{
\begin{tabular}{
    >{\centering\arraybackslash}m{2.0cm}
    >{\centering\arraybackslash}m{3.5cm}
    >{\centering\arraybackslash}m{1.6cm}
    >{\centering\arraybackslash}m{1.4cm}
}
    \rowcolor{gray!20}
    \multicolumn{4}{c}{\textbf{Base Model Table}} \\
    \toprule
    \textbf{Category} & \textbf{Model} & \textbf{w/o} & \textbf{Cot} \\
    \midrule
    Dense Model & Qwen2.5-72B & \textbf{91.02\%} & --- \\
    Dense Model & Qwen2.5-7B few-shot & 87.31\% & 69.88\% \\
    Dense Model & Qwen2.5-0.5B few-shot & 35.29\% & 29.41\% \\
    
    \rowcolor{gray!20}
    \multicolumn{4}{c}{\textbf{Mixed Reasoning Model Table}} \\
    \toprule
    \textbf{Category} & \textbf{Model} & \textbf{w/o} & \textbf{Thinking} \\
    \midrule
    Mixed & Qwen3-235B-a33B few-shot & \textbf{92.88\%} & --- \\
    Mixed & Qwen3-30B-a3B few-shot & 89.47\% & --- \\
    Mixed & Qwen3-32B few-shot & 90.40\% & --- \\
    
    \rowcolor{gray!20}
    \multicolumn{4}{c}{\textbf{Reasoning Model Table}} \\
    \toprule
    \textbf{Category} & \textbf{Model} & \textbf{Reasoning} & --- \\
    \midrule
    Reasoning & Deepseek r1 & \textbf{92.88\%} & --- \\
    Reasoning & Deepseek r1 distill 7B & 40.56\% & --- \\
    Reasoning & Deepseek r1 distill 14B & 81.11\% & --- \\
    Reasoning & Deepseek r1 distill 32B & 86.69\% & --- \\
    Reasoning & QwQ-32B & --- & --- \\
    Reasoning & Kongzi-0.5B & 59.37\% & --- \\
    Reasoning & Kongzi-7B & 87.56\% & --- \\
    Reasoning & Kongzi-0.5B (RL) & 71.21\% & --- \\
    Reasoning & Kongzi-7B (RL) & 90.37\% & --- \\
    \bottomrule
\end{tabular}
}

\vspace{0.1cm}
\captionsetup{justification=raggedright,singlelinecheck=false}
\caption{Results on CMMLU-Chinese-History. Best scores in each metric are highlighted in bold. `w/o` refers to accuracy without enhancement prompts (e.g., Cot or Thinking). `---` indicates the model was not tested or data is unavailable under the corresponding condition.}
\label{tab:cmmlu_kongzi}
\end{table}

%% file: table/table1.tex
\begin{table*}[t]  
\centering
\fontsize{9}{12}\selectfont
\renewcommand{\arraystretch}{1.3}
\setlength{\tabcolsep}{6pt}
\rowcolors{3}{white}{gray!5}

\resizebox{\textwidth}{!}{
\begin{tabular}{
    >{\centering\arraybackslash}m{3.8cm}
    *{5}{>{\centering\arraybackslash}m{1.7cm}}
    >{\centering\arraybackslash}m{2.1cm}
}
\rowcolor{gray!20}
\toprule
\textbf{Model} & 
\makecell{\textbf{Think}\\ \textbf{Score}} & 
\makecell{\textbf{Mean}\\ \textbf{Answer}\\ \textbf{Score}} & 
\makecell{\textbf{Historical}\\ \textbf{Accuracy}} & 
\makecell{\textbf{Logical}\\ \textbf{Reasoning}} & 
\makecell{\textbf{Problem}\\ \textbf{Solving}} & 
\makecell{\small \textbf{$>$Deepseek-r1}\\ \textbf{Ratio}} \\
\toprule

\rowcolor{gray!10}
\multicolumn{7}{l}{\textbf{\textcolor{navy_blue}{GPT-4o}}} \\ 
\midrule
Deepseek-r1-distill-7B & 45.66 & 37.82 & 24.29 & 49.54 & 39.63 & 24.49\% \\
Deepseek-r1-distill-14B & 62.41 & 60.86 & 43.92 & 74.65 & 64.01 & 68.48\% \\
Deepseek-r1-distill-32B & 60.13 & 60.83 & 44.70 & 74.88 & 62.90 & 70.07\% \\
Deepseek-r1 & 67.81 & 59.94 & 46.31 & 70.14 & 63.36 & -- \\
O3-mini & 32.77 & 61.05 & 45.90 & 74.06 & 63.18 & 53.00\% \\
QwQ-32B-preview & -- & 57.97 & 42.53 & 71.84 & 59.54 & 49.77\% \\
\textbf{Kongzi-0.5B (Ours)} & 60.42 & 56.39 & 53.34 & 62.41 & 53.43 & 62.28\% \\
\textbf{Kongzi-7B (Ours)} & \textbf{72.34} & \textbf{70.64} & \textbf{58.45} & \textbf{82.13} & \textbf{71.34} & \textbf{94.32}\% \\

\midrule
\rowcolor{gray!10}
\multicolumn{7}{l}{\textbf{\textcolor{navy_blue}{Gemini-2.5}}} \\ 
\midrule
Deepseek-r1-distill-7B & 62.09 & 54.68 & 45.94 & 60.50 & 57.60 & 22.35\% \\
Deepseek-r1-distill-14B & 75.64 & 78.53 & 70.60 & 82.95 & 82.03 & 44.70\% \\
Deepseek-r1-distill-32B & 75.12 & 76.55 & 68.25 & 81.57 & 79.82 & 41.04\% \\
Deepseek-r1 & 77.39 & 76.05 & 69.30 & 79.95 & 78.89 & -- \\
O3-mini & 33.09 & 84.46 & 77.19 & \textbf{88.99} & 87.19 & 32.03\% \\
QwQ-32B-preview & -- & 77.13 & 69.40 & 82.53 & 79.45 & 27.42\% \\
\textbf{Kongzi-0.5B (Ours)} & 77.91 & 68.49 & 61.13 & 69.88 & 74.45 & 48.17\% \\
\textbf{Kongzi-7B (Ours)} & \textbf{89.24} & \textbf{85.10} & \textbf{80.45} & 85.43 & \textbf{89.42} & \textbf{68.21}\% \\

\midrule
\rowcolor{gray!10}
\multicolumn{7}{l}{\textbf{\textcolor{navy_blue}{Qwen-Max}}} \\ 
\midrule
Deepseek-r1-distill-7B & 44.50 & 34.72 & 19.86 & 50.37 & 33.92 & 14.98\% \\
Deepseek-r1-distill-14B & 59.93 & 57.57 & 39.03 & 78.39 & 55.30 & 65.21\% \\
Deepseek-r1-distill-32B & 61.52 & 58.49 & 39.82 & 79.95 & 55.71 & 62.90\% \\
Deepseek-r1 & 64.29 & 59.21 & 43.01 & 74.70 & 59.91 & -- \\
O3-mini & 32.77 & 61.05 & 45.90 & 74.06 & \textbf{63.18} & 54.38\% \\
QwQ-32B-preview & -- & 57.80 & 40.97 & 77.05 & 55.39 & 50.23\% \\
\textbf{Kongzi-0.5B (Ours)} & 60.40 & 50.49 & 39.82 & 67.12 & 44.52 & 49.18\% \\
\textbf{Kongzi-7B (Ours)} & \textbf{73.31} & \textbf{67.82} & \textbf{56.36} & \textbf{84.78} & 62.31 & \textbf{87.21}\% \\

\bottomrule
\end{tabular}
}

\vspace{0.5em}
\caption{\textbf{Comparison of Kongzi with other models on various metrics.} 
\textit{Think Score} is the rating given by the LLM for the thinking process. 
\textit{Mean Answer Score} is recomputed as the arithmetic mean of the displayed Historical Accuracy, Logical Reasoning, and Problem Solving scores, then rounded to two decimals. It is a derived summary, not an additional evaluation run. 
\textit{$>$Deepseek-r1 Ratio} retains the reported percentage of cases favoring the model over Deepseek-r1; per-item win/tie counts and aggregation details are unavailable in the current materials. 
The best scores in each metric are in \textbf{bold}.}

\label{tab:comparison}
\end{table*}

%% file: sections/conclusion.tex
\section{Conclusion}
We presented Kongzi, a historical language model that combines domain-specific continued pre-training, supervised fine-tuning, and a factuality reward based on knowledge-graph similarity. In the reported evaluations, RL improves the derived Mean Answer Score for both the 0.5B and 7B models under all three LLM judges. Kongzi-7B receives the highest Historical Accuracy ratings among the compared models, while its CMMLU-Chinese-History accuracy remains below the strongest reported baselines. These descriptive findings motivate further study of factuality rewards in historical reasoning. Future work will examine broader historical corpora and evaluation settings, alongside more detailed reporting of judge outputs and aggregation procedures.

\section*{Limitations}
Our experiments cover Chinese history and two model sizes; generality across domains, languages, and model scales remains to be established. The current materials do not provide per-item judge records or win/tie counts, nor specify effective evaluation denominators or aggregation across judging rounds. We therefore report descriptive comparisons and cannot derive reliable confidence intervals or establish statistical significance from these aggregates. Stronger judge ratings do not imply that generated narratives are free of hallucinations or that the factuality reward is universally effective.

\section*{Ethics Statement}
Our work studies factual consistency and reasoning in historical question answering using historical corpora, a constructed test set, and CMMLU-Chinese-History. Historical sources and model outputs may contain biases or inaccurate interpretations. Counterfactual responses are hypothetical analyses and should be distinguished from statements about recorded events. We do not claim that the models' outputs are free of factual errors or harmful content.

%% file: table/table3.tex
\begin{table}[H]  
\fontsize{9}{12}\selectfont
\vspace{+0.2in}
\centering
\renewcommand{\arraystretch}{1.2}
\setlength{\tabcolsep}{7pt}
\rowcolors{4}{white}{gray!5}

\resizebox{\textwidth}{!}{
\begin{tabular}{>{\centering\arraybackslash}m{2.4cm} *{4}{>{\centering\arraybackslash}m{2.1cm}}}
\rowcolor{gray!20}
\toprule
\textbf{Parameter} & \textbf{LR} & \textbf{Steps / Epochs} & \textbf{Batch Size} & \textbf{Seq Length} \\
\midrule

\rowcolor{gray!10}
\multicolumn{5}{l}{\textbf{\textcolor{navy_blue}{1. Continue Pretrain Stage}}} \\
Kongzi-0.5B & 2e-5 & 1500 & 128 & 8192 \\
Kongzi-7B   & 5e-5 & 750  & 256 & 8192 \\
\rowcolor{white}
\textbf{Optimizer} & \multicolumn{2}{c}{AdamW} & \textbf{Hardware} & \multicolumn{1}{c}{NVIDIA A6000} \\
\midrule

\rowcolor{gray!10}
\multicolumn{5}{l}{\textbf{\textcolor{navy_blue}{2. SFT Stage}}} \\
Kongzi-0.5B & 2e-5 & 4000 & 128 & 4096 \\
Kongzi-7B   & 5e-5 & 2000 & 256 & 4096 \\
\rowcolor{white}
\textbf{Optimizer} & \multicolumn{2}{c}{AdamW} & \textbf{Hardware} & \makecell{NVIDIA A6000\\8$\times$RTX 3090} \\
\midrule

\rowcolor{gray!10}
\multicolumn{5}{l}{\textbf{\textcolor{navy_blue}{3. COT-SFT Stage}}} \\
Kongzi-0.5B & 2e-5 & 390  & 128 & 4096 \\
Kongzi-7B   & 5e-5 & 195  & 256 & 4096 \\
\rowcolor{white}
\textbf{Optimizer} & \multicolumn{2}{c}{AdamW} & \textbf{Hardware} & \multicolumn{1}{c}{NVIDIA A6000} \\
\midrule

\rowcolor{gray!10}
\multicolumn{5}{l}{\textbf{\textcolor{navy_blue}{4. COT-RL Stage}}} \\
Kongzi-0.5B & 2e-5 & 117  & 128 & 4096 \\
Kongzi-7B   & 5e-5 & 117  & 256 & 4096 \\
\rowcolor{white}
\textbf{Optimizer} & \multicolumn{2}{c}{AdamW} & \textbf{Hardware} & \multicolumn{1}{c}{NVIDIA A6000} \\


\bottomrule
\end{tabular}
}

\vspace{1em}
\caption{\textbf{Training configuration across different stages.}}  
\label{tab:vertical_config}
\end{table}

%% file: image/kong11.tex
\begin{figure*}[ht]
    \centering
    \includegraphics[width=\textwidth]{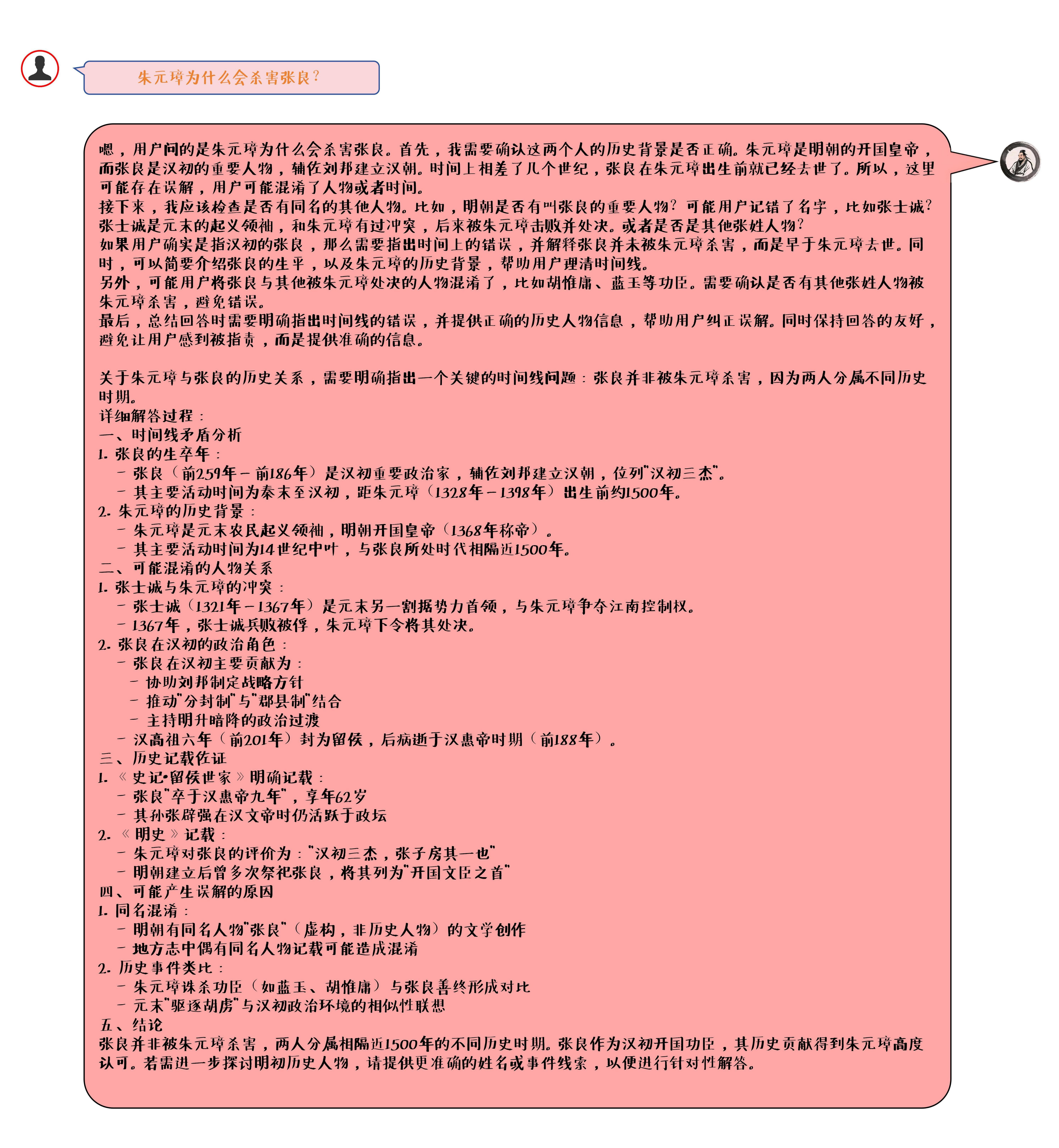}
    \vspace{+0.1in}
    \caption{\textbf{Spatial-temporal relationship comprehension test, the results of Kongzi.}}
    \label{fig:kong11}
\end{figure*}

%% file: image/kong12.tex
\begin{figure*}[ht]
    \centering
    \includegraphics[width=\textwidth]{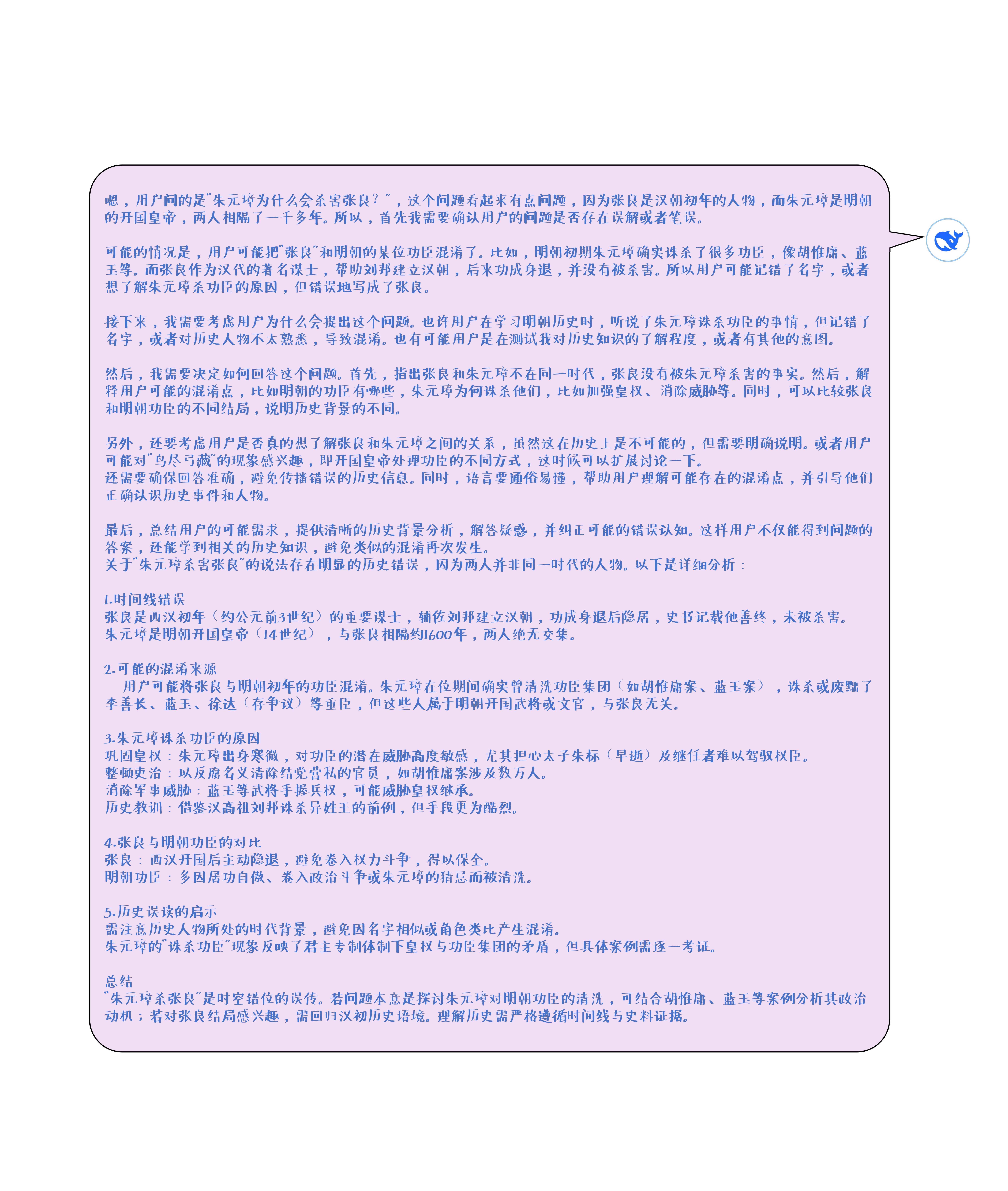}
    \vspace{+0.1in}
    \caption{\textbf{Spatial-temporal relationship comprehension test, the results of Deepseek.}}
    \label{fig:kong12}
\end{figure*}

%% file: image/kong13.tex
\begin{figure*}[ht]
    \centering
    \includegraphics[width=\textwidth]{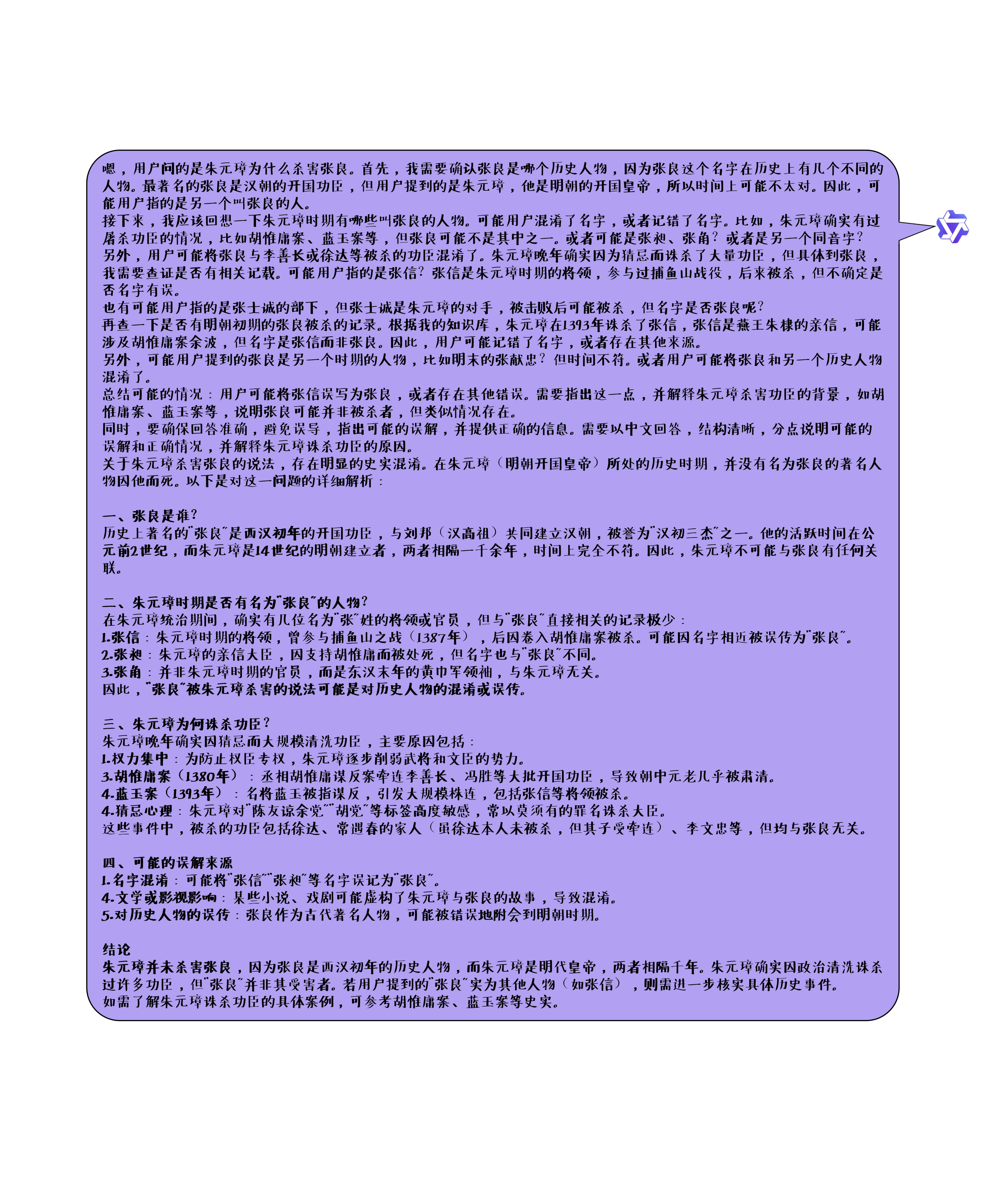}
    \vspace{+0.1in}
    \caption{\textbf{Spatial-temporal relationship comprehension test, the results of QWQ.}}
    \label{fig:kong13}
\end{figure*}

%% file: image/kong21.tex
\begin{figure*}[ht]
    \centering
    \includegraphics[width=\textwidth]{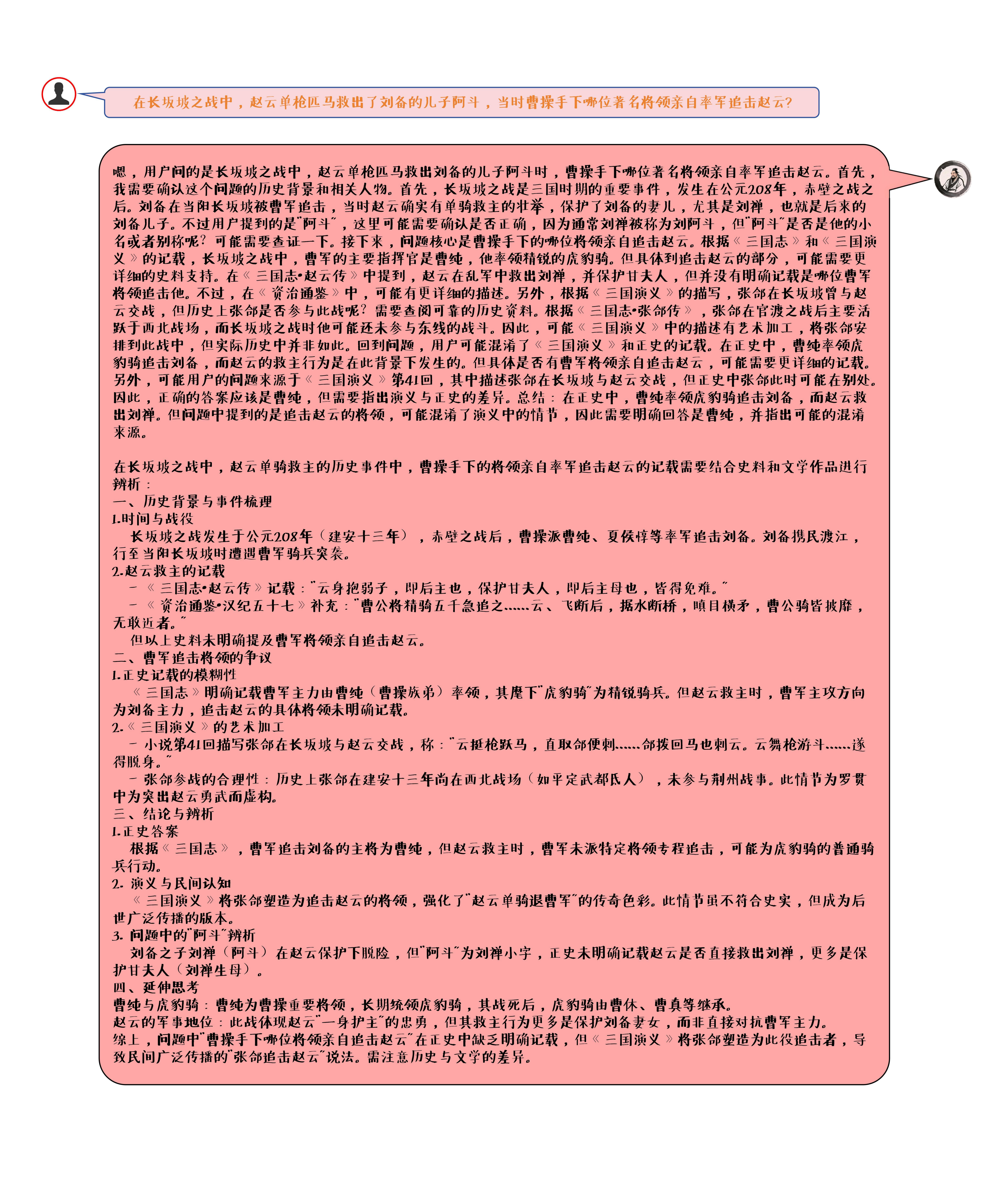}
    \vspace{+0.1in}
    \caption{\textbf{Historical Fact Information Test, the results of Kongzi.}}
    \label{fig:kong21}
\end{figure*}

%% file: image/kong22.tex
\begin{figure*}[ht]
    \centering
    \includegraphics[width=\textwidth]{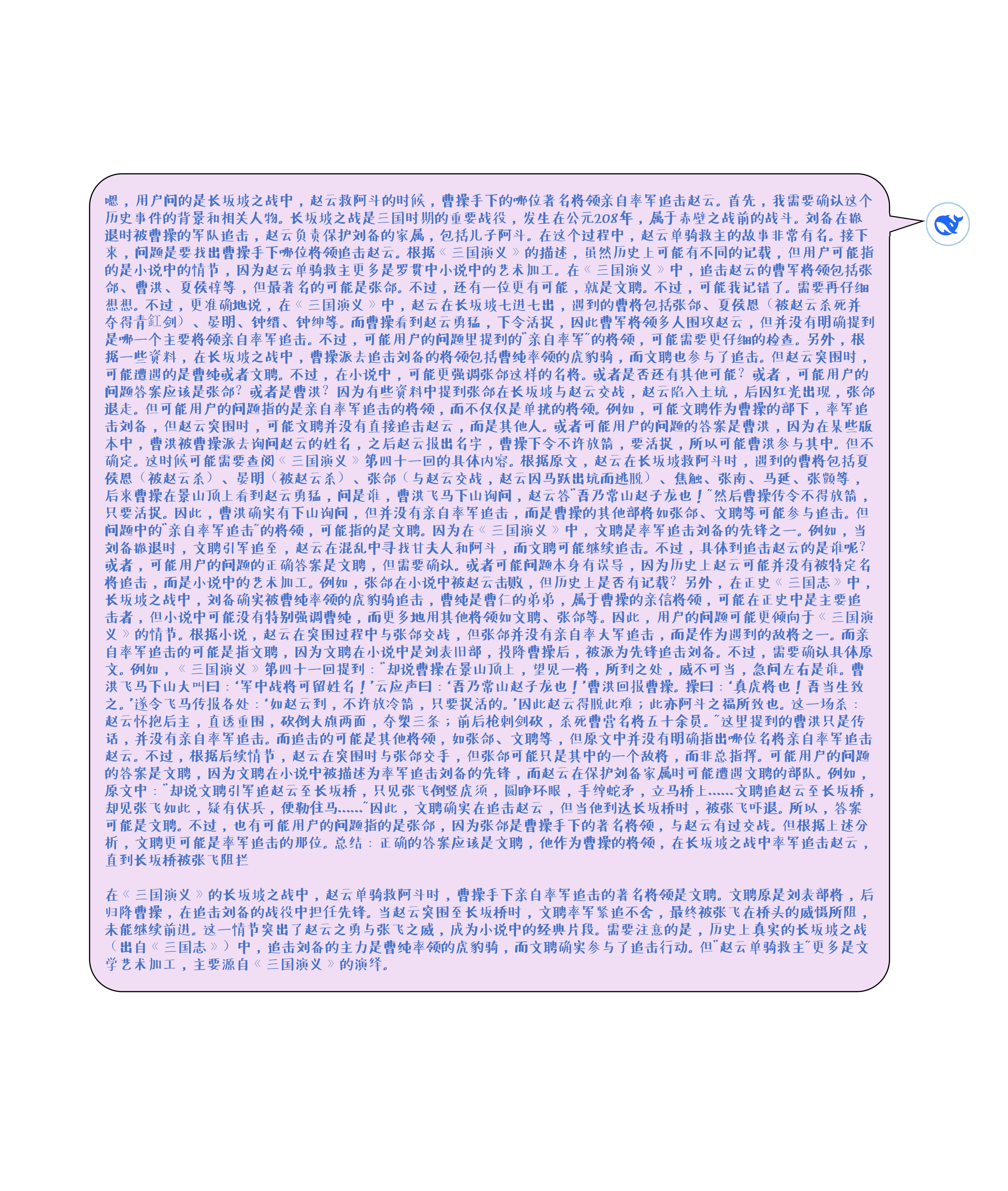}
    \vspace{+0.1in}
    \caption{\textbf{Historical Fact Information Test, the results of Deepseek.}}
    \label{fig:kong22}
\end{figure*}

%% file: image/kong23.tex
\begin{figure*}[ht]
    \centering
    \includegraphics[width=\textwidth]{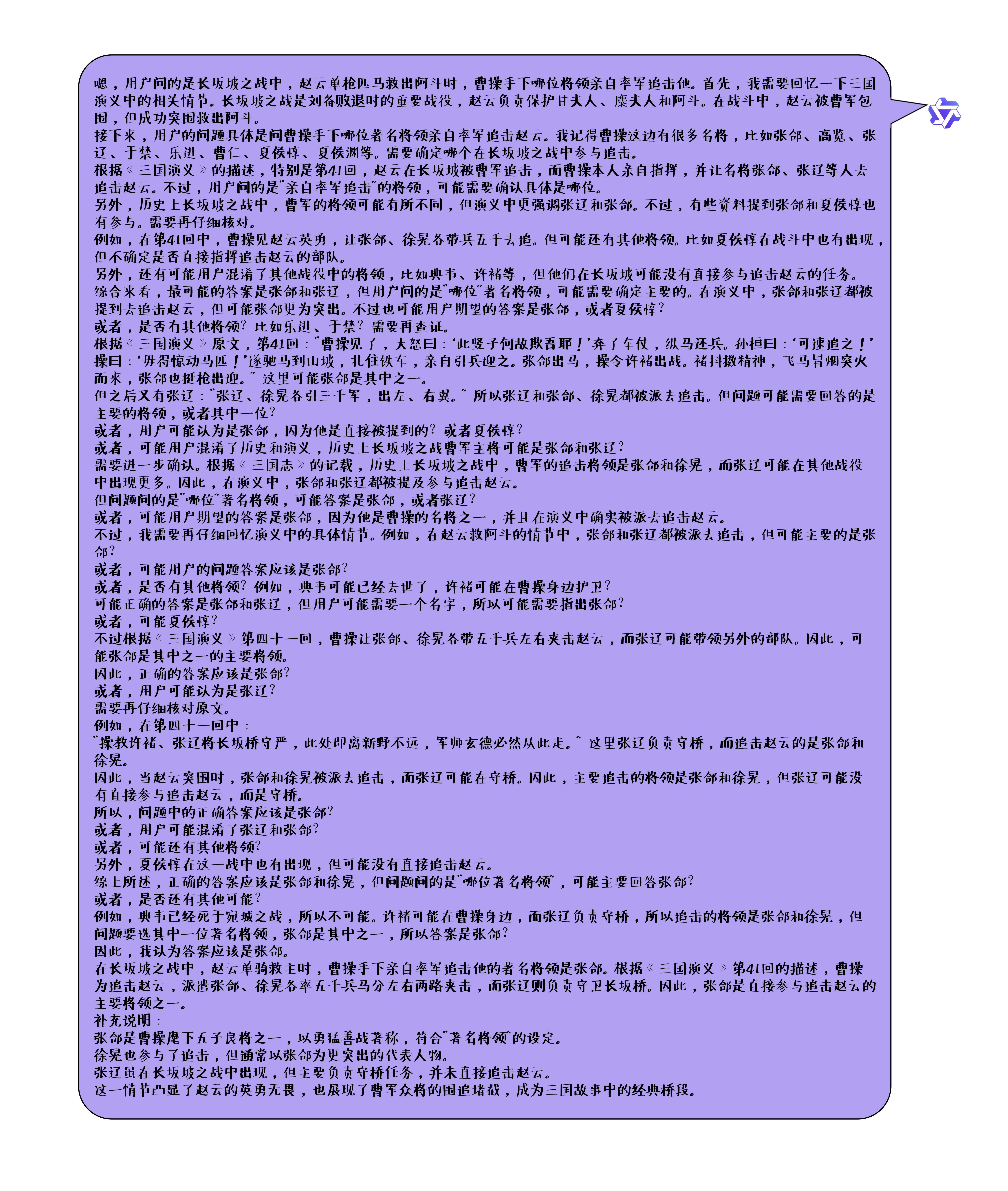}
    \vspace{+0.1in}
    \caption{\textbf{Historical Fact Information Test, the results of QWQ.}}
    \label{fig:kong23}
\end{figure*}

%% file: image/kong30.tex
\begin{figure*}[ht]
    \centering
    \includegraphics[width=\textwidth]{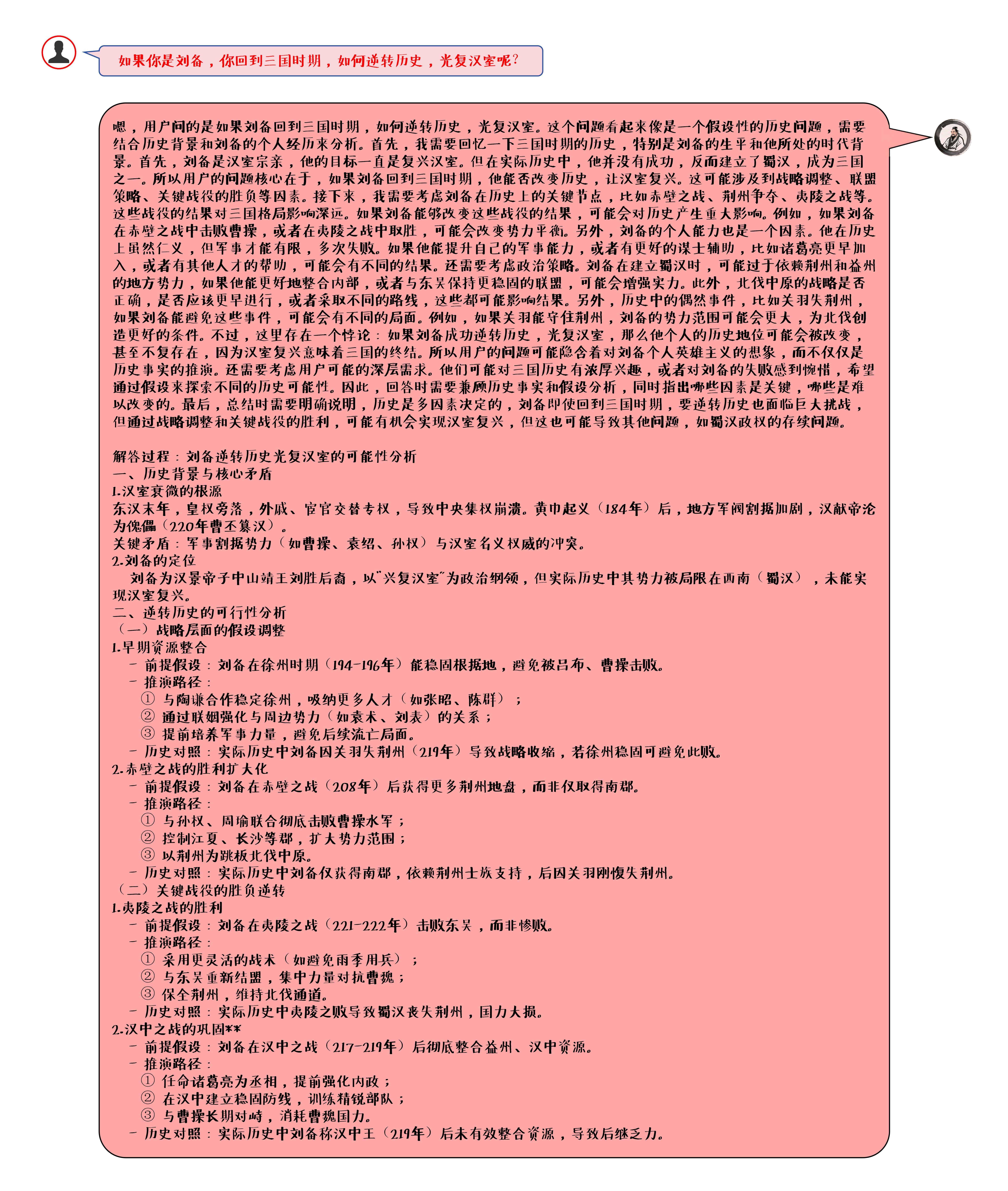}
    \vspace{+0.1in}
    \caption{\textbf{Open-ended Situational Reasoning Q\&A, the results of Kongzi.}}
    \label{fig:kong30}
\end{figure*}

%% file: image/kong31.tex
\begin{figure*}[ht]
    \centering
    \includegraphics[width=\textwidth]{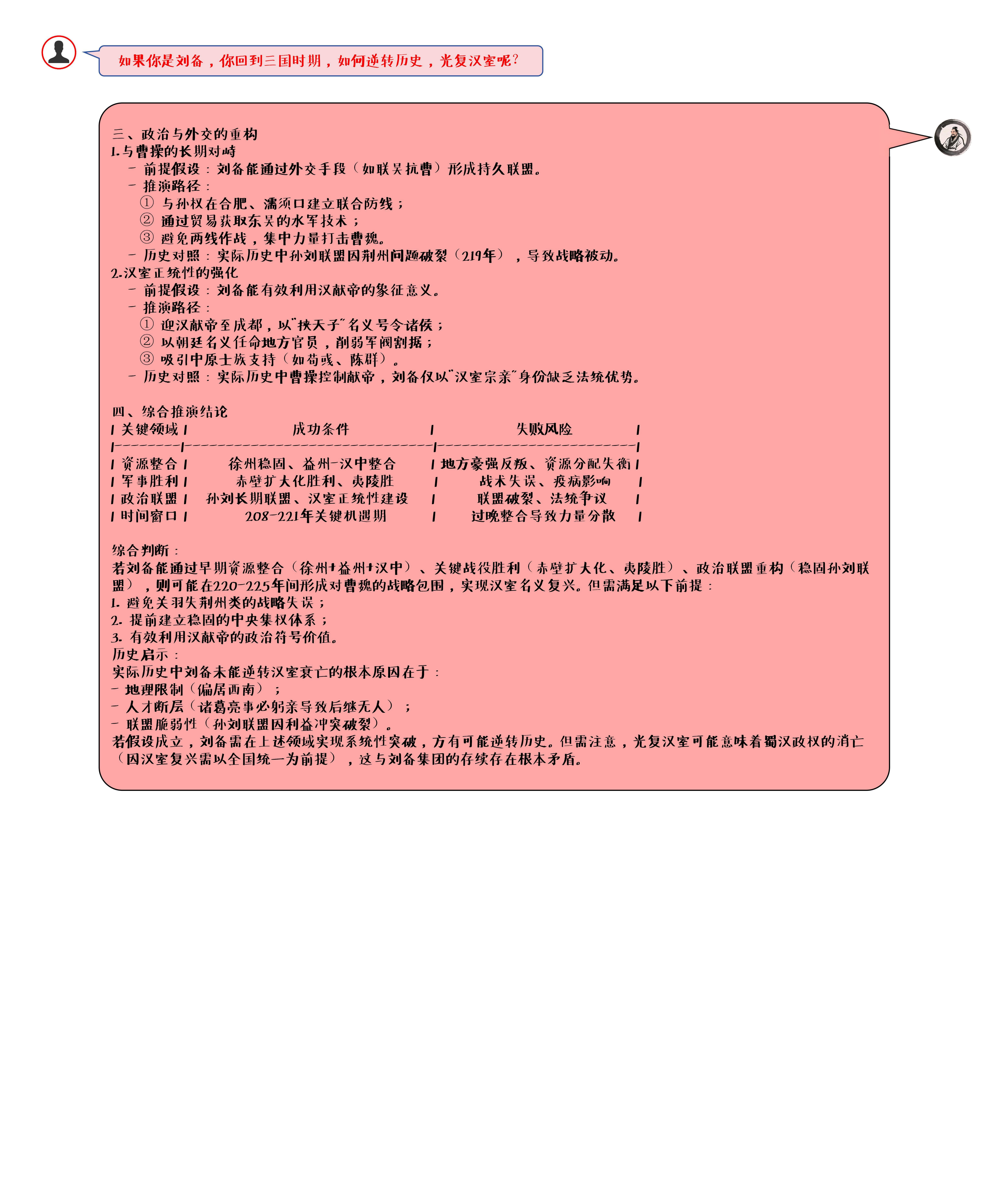}
    \vspace{+0.1in}
    \caption{\textbf{Open-ended Situational Reasoning Q\&A, the results of Kongzi.}}
    \label{fig:kong31}
\end{figure*}

%% file: image/kong32.tex
\begin{figure*}[ht]
    \centering
    \includegraphics[width=\textwidth]{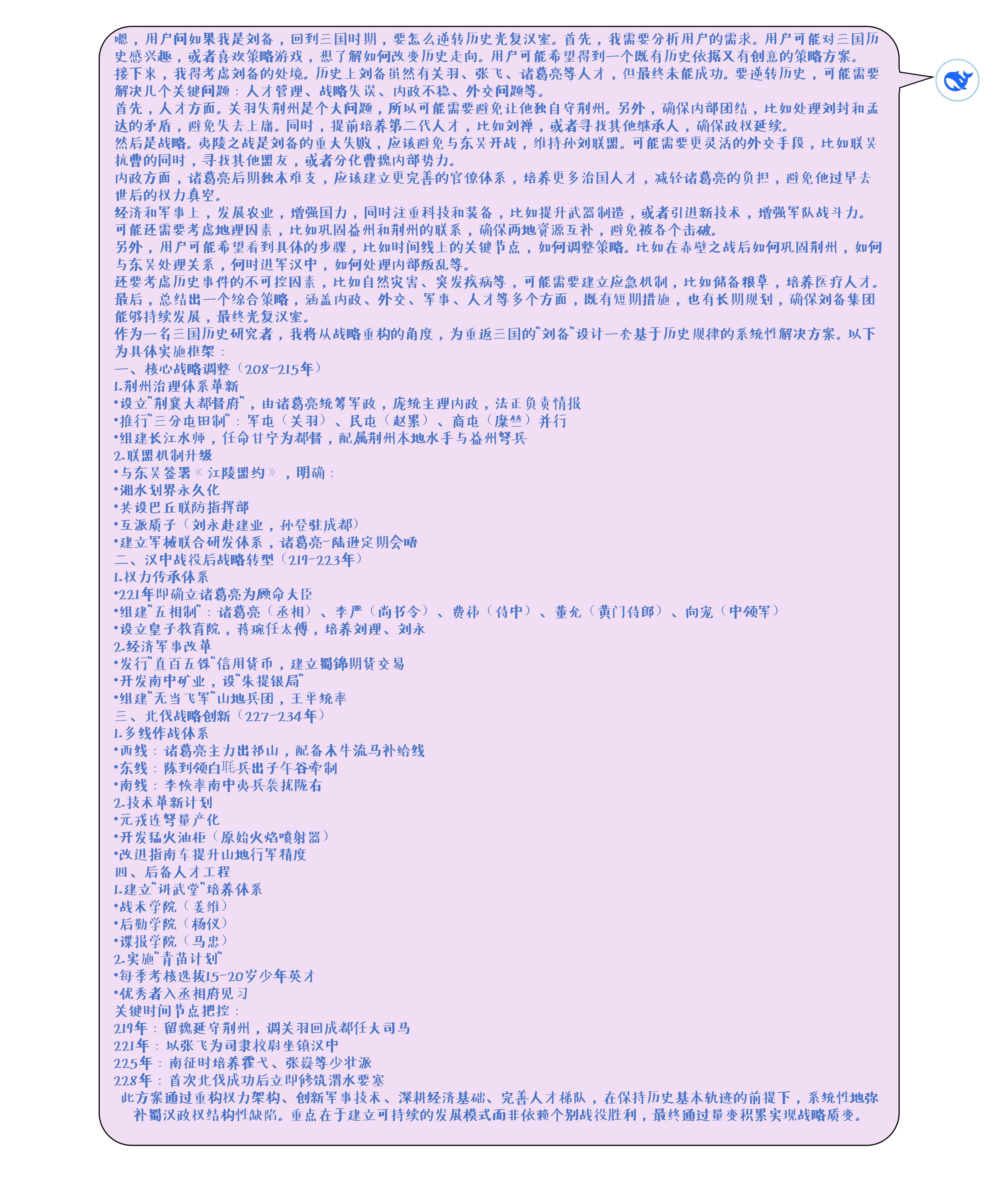}
    \vspace{+0.1in}
    \caption{\textbf{Open-ended Situational Reasoning Q\&A, the results of Deepseek.}}
    \label{fig:kong32}
\end{figure*}

%% file: image/kong33.tex
\begin{figure*}[ht]
    \centering
    \includegraphics[width=\textwidth]{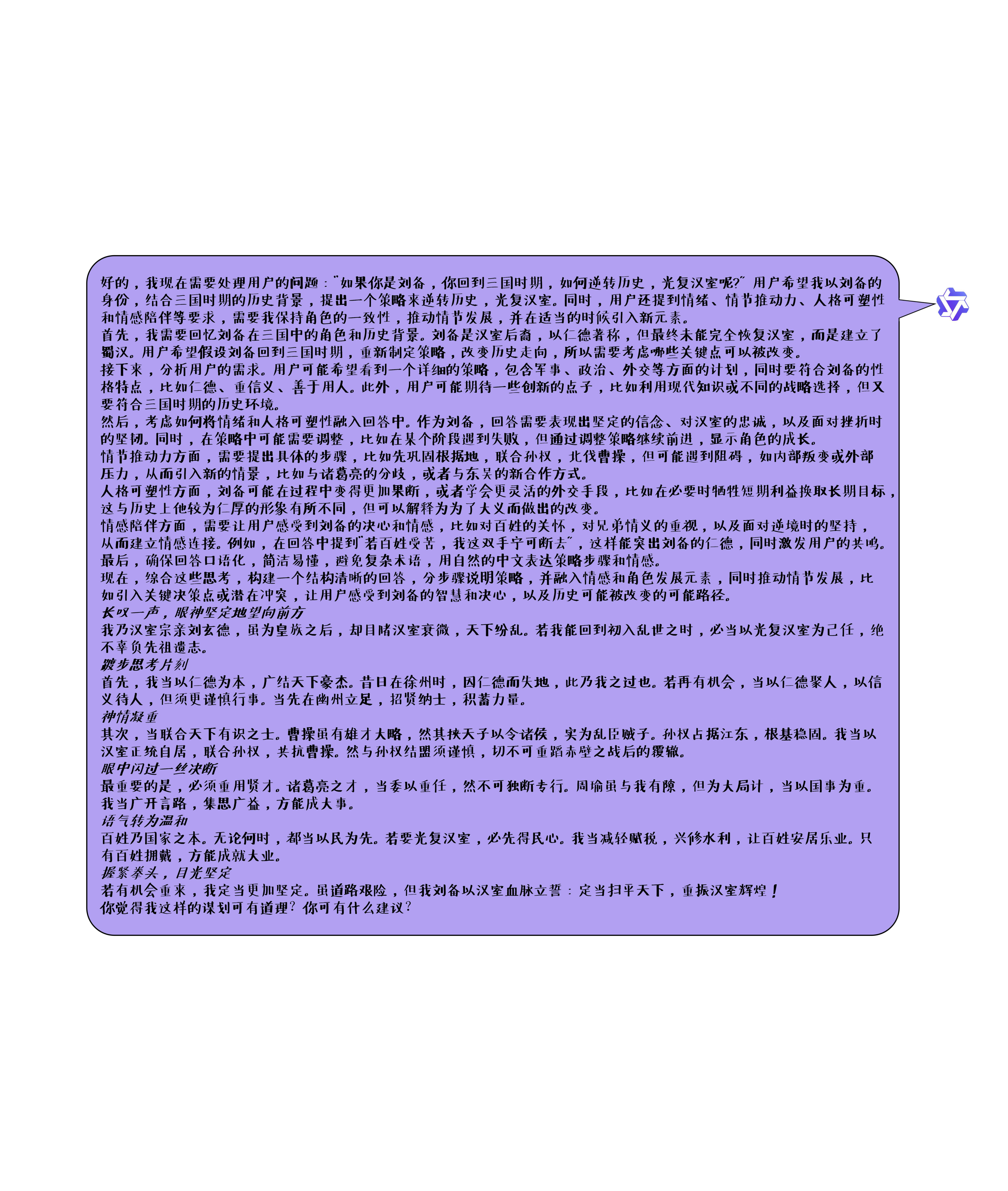}
    \vspace{+0.1in}
    \caption{\textbf{Open-ended Situational Reasoning Q\&A, the results of QWQ.}}
    \label{fig:kong33}
\end{figure*}

%% file: main.bbl
\begin{thebibliography}{23}
\providecommand{\natexlab}[1]{#1}

\bibitem[{Cao et~al.(2024)Cao, Peng, Zhang, Shi, Liu, Ding, and
  Jin}]{cao2024tonggumasteringclassicalchinese}
Jiahuan Cao, Dezhi Peng, Peirong Zhang, Yongxin Shi, Yang Liu, Kai Ding, and
  Lianwen Jin. 2024.
\newblock \href {https://arxiv.org/abs/2407.03937} {Tonggu: Mastering classical
  chinese understanding with knowledge-grounded large language models}.
\newblock \emph{Preprint}, arXiv:2407.03937.

\bibitem[{Chen et~al.(2024{\natexlab{a}})Chen, Liu, Chen, Gu, Wu, Tao, Fu, and
  Ye}]{chen2024inside}
Chao Chen, Kai Liu, Ze~Chen, Yi~Gu, Yue Wu, Mingyuan Tao, Zhihang Fu, and
  Jieping Ye. 2024{\natexlab{a}}.
\newblock \href {https://openreview.net/forum?id=Zj12nzlQbz} {{INSIDE}: {LLM}s'
  internal states retain the power of hallucination detection}.
\newblock In \emph{The Twelfth International Conference on Learning
  Representations}.

\bibitem[{Chen et~al.(2024{\natexlab{b}})Chen, Huang, Gao, Wang, Zhao, and
  Ding}]{chen2024learning}
Xin Chen, Hanxian Huang, Yanjun Gao, Yi~Wang, Jishen Zhao, and Ke~Ding.
  2024{\natexlab{b}}.
\newblock Learning to maximize mutual information for chain-of-thought
  distillation.
\newblock \emph{arXiv preprint arXiv:2403.03348}.

\bibitem[{DeepSeek-AI et~al.(2025)DeepSeek-AI, Guo, Yang, Zhang, Song, Zhang,
  Xu, Zhu, Ma, Wang, Bi, Zhang, Yu, Wu, Wu, Gou, Shao, Li, Gao, Liu, Xue, Wang,
  Wu, Feng, Lu, Zhao, Deng, Zhang, Ruan, Dai, Chen, Ji, Li, Lin, Dai, Luo, Hao,
  Chen, Li, Zhang, Bao, Xu, Wang, Ding, Xin, Gao, Qu, Li, Guo, Li, Wang, Chen,
  Yuan, Qiu, Li, Cai, Ni, Liang, Chen, Dong, Hu, Gao, Guan, Huang, Yu, Wang,
  Zhang, Zhao, Wang, Zhang, Xu, Xia, Zhang, Zhang, Tang, Li, Wang, Li, Tian,
  Huang, Zhang, Wang, Chen, Du, Ge, Zhang, Pan, Wang, Chen, Jin, Chen, Lu,
  Zhou, Chen, Ye, Wang, Yu, Zhou, Pan, Li, Zhou, Wu, Ye, Yun, Pei, Sun, Wang,
  Zeng, Zhao, Liu, Liang, Gao, Yu, Zhang, Xiao, An, Liu, Wang, Chen, Nie,
  Cheng, Liu, Xie, Liu, Yang, Li, Su, Lin, Li, Jin, Shen, Chen, Sun, Wang,
  Song, Zhou, Wang, Shan, Li, Wang, Wei, Zhang, Xu, Li, Zhao, Sun, Wang, Yu,
  Zhang, Shi, Xiong, He, Piao, Wang, Tan, Ma, Liu, Guo, Ou, Wang, Gong, Zou,
  He, Xiong, Luo, You, Liu, Zhou, Zhu, Xu, Huang, Li, Zheng, Zhu, Ma, Tang,
  Zha, Yan, Ren, Ren, Sha, Fu, Xu, Xie, Zhang, Hao, Ma, Yan, Wu, Gu, Zhu, Liu,
  Li, Xie, Song, Pan, Huang, Xu, Zhang, and
  Zhang}]{deepseekai2025deepseekr1incentivizingreasoningcapability}
DeepSeek-AI, Daya Guo, Dejian Yang, Haowei Zhang, Junxiao Song, Ruoyu Zhang,
  Runxin Xu, Qihao Zhu, Shirong Ma, Peiyi Wang, Xiao Bi, Xiaokang Zhang,
  Xingkai Yu, Yu~Wu, Z.~F. Wu, Zhibin Gou, Zhihong Shao, Zhuoshu Li, Ziyi Gao,
  and 181 others. 2025.
\newblock \href {https://arxiv.org/abs/2501.12948} {Deepseek-r1: Incentivizing
  reasoning capability in llms via reinforcement learning}.
\newblock \emph{Preprint}, arXiv:2501.12948.

\bibitem[{Gu et~al.(2025)Gu, Jiang, Shi, Tan, Zhai, Xu, Li, Shen, Ma, Liu,
  Wang, Zhang, Wang, Gao, Ni, and Guo}]{gu2025surveyllmasajudge}
Jiawei Gu, Xuhui Jiang, Zhichao Shi, Hexiang Tan, Xuehao Zhai, Chengjin Xu, Wei
  Li, Yinghan Shen, Shengjie Ma, Honghao Liu, Saizhuo Wang, Kun Zhang, Yuanzhuo
  Wang, Wen Gao, Lionel Ni, and Jian Guo. 2025.
\newblock \href {https://arxiv.org/abs/2411.15594} {A survey on
  llm-as-a-judge}.
\newblock \emph{Preprint}, arXiv:2411.15594.

\bibitem[{Huang et~al.(2025)Huang, Yu, Ma, Zhong, Feng, Wang, Chen, Peng, Feng,
  Qin, and Liu}]{Huang_2025}
Lei Huang, Weijiang Yu, Weitao Ma, Weihong Zhong, Zhangyin Feng, Haotian Wang,
  Qianglong Chen, Weihua Peng, Xiaocheng Feng, Bing Qin, and Ting Liu. 2025.
\newblock \href {https://doi.org/10.1145/3703155} {A survey on hallucination in
  large language models: Principles, taxonomy, challenges, and open questions}.
\newblock \emph{ACM Transactions on Information Systems}, 43(2):1--55.

\bibitem[{Kim et~al.(2023)Kim, Joo, Kim, Jang, Ye, Shin, and Seo}]{kim2023the}
Seungone Kim, Se~June Joo, Doyoung Kim, Joel Jang, Seonghyeon Ye, Jamin Shin,
  and Minjoon Seo. 2023.
\newblock \href {https://openreview.net/forum?id=D7omx8QyFP} {The cot
  collection: Improving zero-shot and few-shot learning of language models via
  chain-of-thought fine-tuning}.
\newblock In \emph{The 2023 Conference on Empirical Methods in Natural Language
  Processing}.

\bibitem[{Li et~al.(2024)Li, Zhang, Li, Peng, Tang, Zhou, Zhang, Hu, Yuan,
  S{\o}gaard, Hershcovich, and Elliott}]{li2024foodieqa}
Wenyan Li, Crystina Zhang, Jiaang Li, Qiwei Peng, Raphael Tang, Li~Zhou, Weijia
  Zhang, Guimin Hu, Yifei Yuan, Anders S{\o}gaard, Daniel Hershcovich, and
  Desmond Elliott. 2024.
\newblock \href {https://doi.org/10.18653/v1/2024.emnlp-main.1063}
  {{F}oodie{QA}: A multimodal dataset for fine-grained understanding of
  {C}hinese food culture}.
\newblock In \emph{Proceedings of the 2024 Conference on Empirical Methods in
  Natural Language Processing}, pages 19077--19095, Miami, Florida, USA.
  Association for Computational Linguistics.

\bibitem[{Nye et~al.(2021)Nye, Andreassen, Gur-Ari, Michalewski, Austin,
  Bieber, Dohan, Lewkowycz, Bosma, Luan, Sutton, and
  Odena}]{nye2021workscratchpadsintermediatecomputation}
Maxwell Nye, Anders~Johan Andreassen, Guy Gur-Ari, Henryk Michalewski, Jacob
  Austin, David Bieber, David Dohan, Aitor Lewkowycz, Maarten Bosma, David
  Luan, Charles Sutton, and Augustus Odena. 2021.
\newblock \href {https://arxiv.org/abs/2112.00114} {Show your work: Scratchpads
  for intermediate computation with language models}.
\newblock \emph{Preprint}, arXiv:2112.00114.

\bibitem[{OpenAI et~al.(2024)OpenAI, Achiam, Adler, Agarwal, Ahmad, Akkaya,
  Aleman, Almeida, Altenschmidt, Altman, Anadkat, Avila, Babuschkin, Balaji,
  Balcom, Baltescu, Bao, Bavarian, Belgum, Bello, Berdine, Bernadett-Shapiro,
  Berner, Bogdonoff, Boiko, Boyd, Brakman, Brockman, Brooks, Brundage, Button,
  Cai, Campbell, Cann, Carey, Carlson, Carmichael, Chan, Chang, Chantzis, Chen,
  Chen, Chen, Chen, Chen, Chess, Cho, Chu, Chung, Cummings, Currier, Dai,
  Decareaux, Degry, Deutsch, Deville, Dhar, Dohan, Dowling, Dunning, Ecoffet,
  Eleti, Eloundou, Farhi, Fedus, Felix, Fishman, Forte, Fulford, Gao, Georges,
  Gibson, Goel, Gogineni, Goh, Gontijo-Lopes, Gordon, Grafstein, Gray, Greene,
  Gross, Gu, Guo, Hallacy, Han, Harris, He, Heaton, Heidecke, Hesse, Hickey,
  Hickey, Hoeschele, Houghton, Hsu, Hu, Hu, Huizinga, Jain, Jain, Jang, Jiang,
  Jiang, Jin, Jin, Jomoto, Jonn, Jun, Kaftan, Łukasz Kaiser, Kamali,
  Kanitscheider, Keskar, Khan, Kilpatrick, Kim, Kim, Kim, Kirchner, Kiros,
  Knight, Kokotajlo, Łukasz Kondraciuk, Kondrich, Konstantinidis, Kosic,
  Krueger, Kuo, Lampe, Lan, Lee, Leike, Leung, Levy, Li, Lim, Lin, Lin, Litwin,
  Lopez, Lowe, Lue, Makanju, Malfacini, Manning, Markov, Markovski, Martin,
  Mayer, Mayne, McGrew, McKinney, McLeavey, McMillan, McNeil, Medina, Mehta,
  Menick, Metz, Mishchenko, Mishkin, Monaco, Morikawa, Mossing, Mu, Murati,
  Murk, Mély, Nair, Nakano, Nayak, Neelakantan, Ngo, Noh, Ouyang, O'Keefe,
  Pachocki, Paino, Palermo, Pantuliano, Parascandolo, Parish, Parparita,
  Passos, Pavlov, Peng, Perelman, de~Avila Belbute~Peres, Petrov,
  de~Oliveira~Pinto, Michael, Pokorny, Pokrass, Pong, Powell, Power, Power,
  Proehl, Puri, Radford, Rae, Ramesh, Raymond, Real, Rimbach, Ross, Rotsted,
  Roussez, Ryder, Saltarelli, Sanders, Santurkar, Sastry, Schmidt, Schnurr,
  Schulman, Selsam, Sheppard, Sherbakov, Shieh, Shoker, Shyam, Sidor, Sigler,
  Simens, Sitkin, Slama, Sohl, Sokolowsky, Song, Staudacher, Such, Summers,
  Sutskever, Tang, Tezak, Thompson, Tillet, Tootoonchian, Tseng, Tuggle,
  Turley, Tworek, Uribe, Vallone, Vijayvergiya, Voss, Wainwright, Wang, Wang,
  Wang, Ward, Wei, Weinmann, Welihinda, Welinder, Weng, Weng, Wiethoff,
  Willner, Winter, Wolrich, Wong, Workman, Wu, Wu, Wu, Xiao, Xu, Yoo, Yu, Yuan,
  Zaremba, Zellers, Zhang, Zhang, Zhao, Zheng, Zhuang, Zhuk, and
  Zoph}]{openai2024gpt4technicalreport}
OpenAI, Josh Achiam, Steven Adler, Sandhini Agarwal, Lama Ahmad, Ilge Akkaya,
  Florencia~Leoni Aleman, Diogo Almeida, Janko Altenschmidt, Sam Altman,
  Shyamal Anadkat, Red Avila, Igor Babuschkin, Suchir Balaji, Valerie Balcom,
  Paul Baltescu, Haiming Bao, Mohammad Bavarian, Jeff Belgum, and 262 others.
  2024.
\newblock \href {https://arxiv.org/abs/2303.08774} {Gpt-4 technical report}.
\newblock \emph{Preprint}, arXiv:2303.08774.

\bibitem[{Ouyang et~al.(2022)Ouyang, Wu, Jiang, Almeida, Wainwright, Mishkin,
  Zhang, Agarwal, Slama, Ray, Schulman, Hilton, Kelton, Miller, Simens, Askell,
  Welinder, Christiano, Leike, and
  Lowe}]{ouyang2022traininglanguagemodelsfollow}
Long Ouyang, Jeff Wu, Xu~Jiang, Diogo Almeida, Carroll~L. Wainwright, Pamela
  Mishkin, Chong Zhang, Sandhini Agarwal, Katarina Slama, Alex Ray, John
  Schulman, Jacob Hilton, Fraser Kelton, Luke Miller, Maddie Simens, Amanda
  Askell, Peter Welinder, Paul Christiano, Jan Leike, and Ryan Lowe. 2022.
\newblock \href {https://arxiv.org/abs/2203.02155} {Training language models to
  follow instructions with human feedback}.
\newblock \emph{Preprint}, arXiv:2203.02155.

\bibitem[{Plaat et~al.(2024)Plaat, Wong, Verberne, Broekens, van Stein, and
  Back}]{plaat2024reasoninglargelanguagemodels}
Aske Plaat, Annie Wong, Suzan Verberne, Joost Broekens, Niki van Stein, and
  Thomas Back. 2024.
\newblock \href {https://arxiv.org/abs/2407.11511} {Reasoning with large
  language models, a survey}.
\newblock \emph{Preprint}, arXiv:2407.11511.

\bibitem[{Qwen et~al.(2025)Qwen, Yang, Yang, Zhang, Hui, Zheng, Yu, Li, Liu,
  Huang, Wei, Lin, Yang, Tu, Zhang, Yang, Yang, Zhou, Lin, Dang, Lu, Bao, Yang,
  Yu, Li, Xue, Zhang, Zhu, Men, Lin, Li, Tang, Xia, Ren, Ren, Fan, Su, Zhang,
  Wan, Liu, Cui, Zhang, and Qiu}]{qwen2025qwen25technicalreport}
Qwen, An~Yang, Baosong Yang, Beichen Zhang, Binyuan Hui, Bo~Zheng, Bowen Yu,
  Chengyuan Li, Dayiheng Liu, Fei Huang, Haoran Wei, Huan Lin, Jian Yang,
  Jianhong Tu, Jianwei Zhang, Jianxin Yang, Jiaxi Yang, Jingren Zhou, Junyang
  Lin, and 24 others. 2025.
\newblock \href {https://arxiv.org/abs/2412.15115} {Qwen2.5 technical report}.
\newblock \emph{Preprint}, arXiv:2412.15115.

\bibitem[{Rafailov et~al.(2024)Rafailov, Sharma, Mitchell, Ermon, Manning, and
  Finn}]{rafailov2024directpreferenceoptimizationlanguage}
Rafael Rafailov, Archit Sharma, Eric Mitchell, Stefano Ermon, Christopher~D.
  Manning, and Chelsea Finn. 2024.
\newblock \href {https://arxiv.org/abs/2305.18290} {Direct preference
  optimization: Your language model is secretly a reward model}.
\newblock \emph{Preprint}, arXiv:2305.18290.

\bibitem[{Ramesh et~al.(2024)Ramesh, Hu, Chaimalas, Mehta, Sessa, Ammar, and
  Bogunovic}]{ramesh2024group}
Shyam~Sundhar Ramesh, Yifan Hu, Iason Chaimalas, Viraj Mehta, Pier~Giuseppe
  Sessa, Haitham~Bou Ammar, and Ilija Bogunovic. 2024.
\newblock \href {https://openreview.net/forum?id=PRAsjrmXXK} {Group robust
  preference optimization in reward-free {RLHF}}.
\newblock In \emph{The Thirty-eighth Annual Conference on Neural Information
  Processing Systems}.

\bibitem[{Schulman et~al.(2017)Schulman, Wolski, Dhariwal, Radford, and
  Klimov}]{schulman2017proximalpolicyoptimizationalgorithms}
John Schulman, Filip Wolski, Prafulla Dhariwal, Alec Radford, and Oleg Klimov.
  2017.
\newblock \href {https://arxiv.org/abs/1707.06347} {Proximal policy
  optimization algorithms}.
\newblock \emph{Preprint}, arXiv:1707.06347.

\bibitem[{Wadhwa et~al.(2024)Wadhwa, Amir, and
  Wallace}]{wadhwa2024investigatingmysteriescotaugmenteddistillation}
Somin Wadhwa, Silvio Amir, and Byron~C. Wallace. 2024.
\newblock \href {https://arxiv.org/abs/2406.14511} {Investigating mysteries of
  cot-augmented distillation}.
\newblock \emph{Preprint}, arXiv:2406.14511.

\bibitem[{Wang et~al.(2023)Wang, Wei, Schuurmans, Le, Chi, Narang, Chowdhery,
  and Zhou}]{wang2023selfconsistencyimproveschainthought}
Xuezhi Wang, Jason Wei, Dale Schuurmans, Quoc Le, Ed~Chi, Sharan Narang,
  Aakanksha Chowdhery, and Denny Zhou. 2023.
\newblock \href {https://arxiv.org/abs/2203.11171} {Self-consistency improves
  chain of thought reasoning in language models}.
\newblock \emph{Preprint}, arXiv:2203.11171.

\bibitem[{Wang et~al.(2026)Wang, Wu, Li, Jiang, and
  Zhou}]{wang2025benchmarkingtemporalreasoningalignment}
Zhenglin Wang, Jialong Wu, Pengfei Li, Yong Jiang, and Deyu Zhou. 2026.
\newblock \href {https://doi.org/10.18653/v1/2026.eacl-short.6} {Benchmarking
  temporal reasoning and alignment across {C}hinese dynasties}.
\newblock In \emph{Proceedings of the 19th Conference of the {E}uropean Chapter
  of the {A}ssociation for {C}omputational {L}inguistics (Volume 2: Short
  Papers)}, pages 105--120, Rabat, Morocco. Association for Computational
  Linguistics.

\bibitem[{Yang et~al.(2025)Yang, Yu, Cui, and
  Wang}]{yang2025reasonfluxhierarchicalllmreasoning}
Ling Yang, Zhaochen Yu, Bin Cui, and Mengdi Wang. 2025.
\newblock \href {https://arxiv.org/abs/2502.06772} {Reasonflux: Hierarchical
  llm reasoning via scaling thought templates}.
\newblock \emph{Preprint}, arXiv:2502.06772.

\bibitem[{Yu et~al.(2023)Yu, He, Wu, Dai, and
  Chen}]{yu2023betterchainofthoughtpromptingstrategies}
Zihan Yu, Liang He, Zhen Wu, Xinyu Dai, and Jiajun Chen. 2023.
\newblock \href {https://arxiv.org/abs/2310.04959} {Towards better
  chain-of-thought prompting strategies: A survey}.
\newblock \emph{Preprint}, arXiv:2310.04959.

\bibitem[{Zhao et~al.(2025)Zhao, Zhou, Ren, Chen, Jia, Zhe, Long, Liu, and
  Lan}]{zhao2025fuxibenchmarkevaluatinglanguage}
Shangqing Zhao, Yuhao Zhou, Yupei Ren, Zhe Chen, Chenghao Jia, Fang Zhe,
  Zhaogaung Long, Shu Liu, and Man Lan. 2025.
\newblock \href {https://doi.org/10.48550/arXiv.2503.15837} {{F\`ux\`i}: A
  benchmark for evaluating language models on ancient chinese text
  understanding and generation}.
\newblock \emph{Preprint}, arXiv:2503.15837.

\bibitem[{Zhong et~al.(2025)Zhong, Shan, Feng, Xiong, Cheng, Zhao, He, Bian,
  and Wang}]{zhong2025dpomeetspporeinforced}
Han Zhong, Zikang Shan, Guhao Feng, Wei Xiong, Xinle Cheng, Li~Zhao, Di~He,
  Jiang Bian, and Liwei Wang. 2025.
\newblock \href {https://arxiv.org/abs/2404.18922} {Dpo meets ppo: Reinforced
  token optimization for rlhf}.
\newblock \emph{Preprint}, arXiv:2404.18922.

\end{thebibliography}
